\documentclass{article}

\usepackage{microtype}
\usepackage{graphicx}
\usepackage{subcaption}
\usepackage{booktabs} 
\usepackage{placeins} 

\usepackage{hyperref}

\usepackage[preprint]{icml2026}

\usepackage{amsmath}
\usepackage{amssymb}
\usepackage{mathtools}
\usepackage{amsthm}

\usepackage[capitalize,noabbrev]{cleveref}

\theoremstyle{plain}
\newtheorem{theorem}{Theorem}[section]

\theoremstyle{definition}
\newtheorem{definition}[theorem]{Definition}

\theoremstyle{remark}

\newcommand{\V}{\mathcal{V}}

\newcommand{\con}{\boldsymbol{c}}

\usepackage[textsize=tiny]{todonotes}

\icmltitlerunning{Dynamic Delayed Tree Expansion For Improved Multi-Path Speculative Decoding}

\begin{document}

\twocolumn[
  \icmltitle{Dynamic Delayed Tree Expansion \\ For Improved Multi-Path Speculative Decoding}



  \icmlsetsymbol{equal}{*}

  \begin{icmlauthorlist}
    \icmlauthor{Rahul Thomas}{rit,col,equal}
    \icmlauthor{Teo Kitanovski}{rit,van,equal}
    \icmlauthor{Micah Goldblum}{rit,col}
    \icmlauthor{Arka Pal}{rit}
  \end{icmlauthorlist}

  \icmlaffiliation{rit}{Ritual}
  \icmlaffiliation{col}{Columbia University}
  \icmlaffiliation{van}{Vanderbilt University}

  \icmlcorrespondingauthor{Rahul Thomas}{rkt2129@columbia.edu}

  \icmlkeywords{Machine Learning, ICML}

  \vskip 0.3in
]



\printAffiliationsAndNotice{\icmlEqualContribution}

\begin{abstract}
  Multi-path speculative decoding accelerates lossless sampling from a target model by using a cheaper draft model to generate a draft tree of tokens, and then applies a verification algorithm that accepts a subset of these. While prior work has proposed various verification algorithms for i.i.d rollouts, their relative performance under matched settings remains unclear. In this work, we firstly present a systematic evaluation of verification strategies across model families, tasks, and sampling regimes, and find that Traversal Verification dominates consistently, with OT-based methods lagging far behind. Our analysis uncovers that this occurs because OT-based methods achieve high multi-token acceptance near the root of the draft tree, while multi-token gains are most impactful deeper in the draft tree, where draft and target distributions diverge. Based on this insight, we propose \textbf{delayed tree expansion}, which drafts a partial single path, delaying the i.i.d. branching point. We show that delayed tree expansion preserves the target distribution and improves on root-node i.i.d rollouts. Further, we develop a dynamic neural selector that estimates the expected block efficiency of optimal-transport-based verification methods from draft and target features, enabling context-dependent expansion decisions. Our neural selector allows OT-based methods like SpecInfer to outperform Traversal Verification for the first time, achieving 5\% higher average throughput across a wide range of models, datasets, and sampling settings.
\end{abstract}

\section{Introduction}

In recent years, large language models (LLMs) have demonstrated impressive performance across a wide range of domains, including but not limited to translation, reasoning, mathematics, and coding \citep{zhu2024multilingualmachinetranslationlarge, kasneci2023chatgptgoodeducation, hendrycks2021math, chen2021evaluating}. Despite such rapid progress, end-to-end latency remains a significant bottleneck in real LLM deployments. Most LLM families, such as Gemma \citep{team2024gemma,team2025gemma}, Qwen \citep{bai2023qwen,hui2024qwen2}, and Llama \citep{touvron2023llama,touvron2023llama2,grattafiori2024llama}, are autoregressive and require an expensive forward pass for each output token. 

Exploiting the fact that the forward pass is memory-bound \citep{fu2024break}, speculative decoding \citep{chen2023accelerating,leviathan2023fast} improves GPU utilization by decoding multiple tokens per pass in a three-step process. First, during \textbf{drafting}, a cheap draft model proposes multiple future tokens. Then, the original LLM performs a \textbf{target} forward pass over these tokens in parallel. Finally, during \textbf{verification}, some tokens are rejected to maintain the target distribution. Throughput gains depend on the drafting and target pass walltimes, as well as block efficiency -- the number of average accepted tokens per target model call.

Recent work has generalized speculative decoding to the \textbf{multi-path} setting, where the draft model proposes a tree of tokens and the target pass uses a tree attention mask. Multi-path methods improve block efficiency by diversifying the draft token pool. In order to maintain the target distribution, specific multi-path verification algorithms are required. These are usually based on optimal transport (OT) solvers that traverse the draft tree in a top-down manner \citep{sun2023spectr,khisti2025multi,miao2024specinfer}, although some incorporate bottom-up traversal \citep{sun2024block,weng2025traversal}. Concurrently, there have been efforts to improve the efficacy of \textbf{drafting} for a fixed verification method via confidence-guided tree construction, adaptive drafting depth, and pipelining \citep{li2024eagle2,brown2024dynamic,xiong2025dyspec,chen2024sequoia,wang2025opt,huang2024specdec++,liu2024pearl,guan2025yggdrasil}.

However, despite the significant body of prior work on multi-path speculative decoding, there has been no systematic and controlled comparison of multi-path verification methods. Most papers introducing new algorithms test against only a handful of relevant baselines, or compare only to single-path drafts. Moreover, prior works which introduce new tree construction methods \citep{li2024eagle2,brown2024dynamic,xiong2025dyspec,chen2024sequoia,wang2025opt,huang2024specdec++,liu2024pearl,guan2025yggdrasil} can often only be used with specific verification algorithms in order to maintain the target distribution. Therefore, an important open question remains: \textbf{which multi-path verification method performs best, and under what conditions?}  Prior works often compare such methods under different drafting methods, datasets, and model families, or only consider the most basic verification methods, making it difficult to measure the true progress of verification gains. 

In this work, we compare multi-path verification methods under matched i.i.d. draft settings across diverse model families, tasks, and sampling regimes. We find that Traversal Verification consistently dominates all other methods. Our analysis reveals that this is the result of wasteful expansion. These other methods, which are all OT-based, achieve highest token acceptances near the root of the draft tree, but the most promising improvements occur deeper in the tree when draft and target distributions diverge. Generating a draft tree from i.i.d. root rollouts forces token diversity into shallow nodes, even when it is more beneficial at deeper nodes.

This motivates us to introduce \textbf{delayed expansion}, where we draft a partial single path, and then expand into i.i.d. rollouts at a "branching point". This drafting policy allows explicit control of three parameters: the single path depth, the expansion factor, and the expansion depth. Optimizing these is a careful balancing act between block efficiency and tree size, because larger trees accept more tokens on average but also incur higher drafting and target pass costs. Following adaptive drafting methods that maximize expected block efficiency offline \citep{chen2024sequoia,brown2024dynamic}, we train a lightweight MLP to predict the optimal values of these parameters from root node features and system-specific draft and target pass times. With this neural predictor, OT-based methods can finally outperform Traversal Verification. In summary, our contributions are as follows:

\begin{itemize}
    \item We provide the first systematic comparison of existing i.i.d. multi-path speculative decoding verification methods across varying model families, datasets, and sampling settings. We find Traversal Verification performs far better than all other methods (OT-based).
    \item To explain the pitfalls of OT-based methods, we analyze acceptance rates and find that the marginal performance improvements between these methods are caused by increasingly diverging target and draft distributions deeper in the tree. Branching early in the tree, where target and draft distributions are most similar, yields relatively less useful acceptance improvements.
    \item To mitigate this, we utilize delayed expansion, where the draft tree is formed by drafting a path and then branching into many i.i.d. paths. We train a context-dependent neural predictor on offline block efficiency estimates to predict optimal path and branch parameters from model and latency features. In online deployments, our method can be applied to all OT-based i.i.d. multi-path methods. Across a multitude of settings, our neural predictor with the OT-based method SpecInfer improves average throughput over Traversal Verification by $\sim 5\%$.
\end{itemize}

\section{Background}
\label{sec:background}

Speculative decoding uses a cheap draft model to decode multiple tokens in one target model call, in three steps:
\begin{enumerate}
    \item \textbf{Drafting.} Using only the draft model, we expand a tree of tokens from the current context and compute next-token draft distributions on the tree nodes.
    \item \textbf{Target Pass.} The target model performs a batched forward pass over the draft tree, with a custom attention mask that respects ancestor-only dependencies, to obtain next-token target distributions on tree nodes.
    \item \textbf{Verification.} Based on the target and draft distributions, we randomly accept a single node on the draft tree and append an additional correction token, such that the output matches the target model distribution.  
\end{enumerate}
The \textbf{acceptance length} $\tau$ is the depth of the accepted node, which grows as the target and draft distributions coincide. The \textbf{block efficiency} $\mathbb{E}[\tau+1]$ is the average number of decoded tokens per target call. When drafting is cheap and the batched target pass takes as long as a target forward pass, block efficiency accurately represents speedup. 

\subsection{Performance Comparisons}

The literature lacks comprehensive comparisons of speculative decoding algorithms. \citet{xia2024unlocking} surveys drafting and verification strategies in speculative decoding, and releases Spec-Bench, a third-party testing environment used to compare SOTA methods under common environment setups. However, their dataset is relatively small, they only support a few verification algorithms, and their evaluations do not incorporate temperature or nucleus sampling variations. More recently, \citet{liu2025speculative} gives the first production-grade vLLM study \citep{kwon2023efficient} on the effects of batch size, model family, and workloads on speculative decoding and variants like EAGLE, EAGLE-3, and multi-token prediction. Still, they use only basic verification algorithms and greedy decoding. We are the first to compare \textit{all} verification algorithms across models, tasks, and sampling settings.

\subsection{Improving Drafting}

Existing work has focused on improving block efficiency by modifying drafting or verification. Verification works build novel algorithms that alter the node selection scheme to improve acceptance lengths. Note that verification and draft improvements can be integrated together when compatible, and many of these drafting methods rely on specific verification methods like NSS \citep{miao2024specinfer}, Naive \citep{chen2023accelerating,leviathan2023fast}, and SpecInfer \citep{miao2024specinfer}. We discuss these in \cref{sec:verification} and focus on drafting here. We focus on works that alter the draft tree construction itself, as these are most relevant to our work. 
We provide further details on orthogonal work in \cref{appendix:orthogonal-work}.

\paragraph{Context-dependent tree structures.} Various works alter the draft topology \textit{online} to only expand nodes with meaningful acceptance improvements. EAGLE-2 \citep{li2024eagle2} deterministically expands top-$k$ tokens from nodes with the highest global draft probabilities, using draft confidence as a proxy for acceptance. Dynamic Depth Decoding \citep{brown2024dynamic} improves EAGLE-2 by adaptively selecting how deep to expand within its deterministic tree structure. EDD and PCT \citep{zheng2025faster} prune candidate branches with low draft confidence, so that the target model compute is spent on branches likely to be accepted. Instead of fixing a deterministic draft tree, DySpec \citep{xiong2025dyspec} dynamically allocates a token tree expansion budget and then samples a draft tree from the token tree online. Our methods also utilize context-dependent tree topologies, but they can more effectively push towards high acceptance regions by relying on offline block efficiency supervision rather than simple proxy metrics.

\paragraph{Offline tree optimization for block efficiency.} Another line of work uses explicit objectives for expected accepted length to select a set of optimal trees offline. Sequoia \citep{chen2024sequoia} uses dynamic programming to select deterministic draft structures under various budget constraints, which optimize SpecInfer block efficiency for sampling without replacement. OPT-Tree \citep{wang2025opt} instead maximizes expected acceptance under NSS. While these methods are similar to ours in that they optimize expected block efficiency, their fixed offline tree collection remains highly sensitive to the dataset and sampling settings, a pitfall which our per-context neural selector naturally overcomes. 

\paragraph{Dynamic draft length control.} Because the search over tree topologies is complex, many draft tree selectors decide how far to draft rather than changing branching logic.  SpecDec++ \citep{huang2024specdec++} learns an adaptive stopping policy to decide when further draft tokens bring minimal benefits. SVIP \citep{zhang2024draft} uses a lightweight draft confidence rule to terminate speculation early when rejections seem likely, to avoid wasted target forward pass work. FailFast \citep{pan2025fail} extends this by using dLLMs for rapid parallelized draft generation, and also rapidly increases draft length in high-acceptance regions to accept massive numbers of tokens. AdaSD \citep{lu2025adasd} adapts draft length by using the divergence between target and draft distributions as a signal to reduce manual tuning. 

\paragraph{Training-based tree policy.} Draft model language modeling performance is not highly correlated with speculative decoding performance \citep{yan2025decoding}. Thus, some works train draft models on objectives more aligned with speculative decoding acceptance. The earliest line of work in this regard, DistillSpec \citep{zhou2023distillspec}, uses knowledge distillation to train the draft model to better align with the target model for Naive acceptance. Online speculative decoding \citep{liu2023online} uses knowledge distillation to update the draft model concurrently with speculative decoding inference, dynamically clustering queries by common domain or topic to improve alignment. AdaSpec \citep{hu2025adaspec} adapts speculative decoding to latency constraints in model serving, prioritizing response time over block efficiency. Group Tree Optimization \citep{hu2025bridging} enhances draft distillation by better aligning the single-path distillation objective with test-time tree-based decoding policies, using a draft tree reward based on expected NSS acceptance and a PPO-style objective that contrasts frozen and evolving draft trees. Our work differs from these in that it expands training specific to NSS and Naive objectives to more diverse verification methods, and in that we learn a compact tree selector model instead of finetuning a large draft model.

\paragraph{Hardware-Aware Tree Decoding.} Some works accelerate the batched forward pass or pick draft tree shapes that are highly compatible with compilers. Sequoia \citep{chen2024sequoia} uses hardware-dependent target and draft times to inform its tree depth and branch selections. Yggdrasil \citep{guan2025yggdrasil} uses an equal-growth tree drafting algorithm to select tree width and depth and verification width to optimize latency and graph compiler optimizations, and improves speculative decoding stage scheduling for minimal GPU to CPU transfer. DeFT \citep{yao2024deft} uses high-utilization QKV grouping and custom kernels to accelerate tree decoding, which is highly relevant to the drafting phase. While we train our neural selector using draft and target pass times, and are hardware-aware like these works, our approach remains highly effective because it incorporates a diversity of verification objectives that these works do not.

\section{Verification Algorithms}
\label{sec:verification}

We now review prior verification algorithms. For the rest of paper, we denote the target model by $M_p$ and the draft model by $M_q$, which share a vocabulary $\V$. We use $p(\cdot|\con)$ and $q(\cdot|\con)$ to denote their next-token distributions over $\V$. All of these algorithms require a formal notion of a draft tree, which we give below. Here, $(\con_1,\con_2)$ denotes the concatenation of contexts. We ignore the root context $\con$ in our $p,q$ and node notation when it is implicitly clear.

\begin{definition}
    A \textbf{draft tree} rooted at $\con$ is a directed tree of nodes represented by \textit{distinct contexts} $\con'$ with root node $\con$, such that for each parent and child node pair $(\con_p,\con_c)$, we have $\con_c=(\con_p,t)$ for some token $t \in \V$. We denote the list of child nodes by $\text{ch}(\con)$, which can duplicate nodes\footnote{This does not follow the standard definition of a child node list in a tree, as child nodes have multiplicity and their order matters.}.
    \label{def:tree}
\end{definition}

For a verification algorithm to preserve the target distribution, it must be compatible with drafting. We cover those which apply to single-path, multi-path, and other drafting regimes. Single-path means the tree is a path sampled from $M_q$, and multi-path is the union of i.i.d. single-paths. 

\begin{table}[h]
\centering
\footnotesize
\setlength\tabcolsep{4pt}
\begin{tabular}{llll}
\toprule
\textbf{Algorithm} & \textbf{Appears In} & \textbf{Multi-Path} & \textbf{OT-Based} \\
\midrule
NaiveTree     & \citet{leviathan2023fast}   & \checkmark & \checkmark \\
NSS       & \citet{miao2024specinfer} &  \checkmark & \checkmark \\
SpecTr    & \citet{sun2023spectr}                     & \checkmark  & \checkmark \\
SpecInfer & \citet{miao2024specinfer}                  & \checkmark  & \checkmark \\
Khisti    & \citet{khisti2025multi}    & \checkmark  & \checkmark \\
Naive     & \citet{chen2023accelerating}      &  &  \\
TV        & \citet{hu2024accelerated}   &  &   \\
BV        & \citet{sun2024block}        &  &   \\
Traversal & \citet{weng2025traversal}     & \checkmark  &   \\
\bottomrule
\end{tabular}
\caption{Summary of all single-path and multi-path verification algorithms, and whether multi-path algorithms are OT-based.}
\label{tab:verification_methods}
\end{table}

\subsection{Single-Path Algorithms}
\label{subsec:single-path}

To the best of our knowledge, there are only three unique\footnote{All multi-path algorithms apply here, but all OT-based ones except NSS degenerate to Naive, and Traversal degenerates to BV. NSS degenerates to a simple algorithm: sample from $p$ until the trajectory no longer matches the draft block.} verification algorithms that apply when the draft tree is an autoregressively sampled length $L$ block $a_{1:L} \sim q(\cdot|\con)$.

\paragraph{Naive speculative sampling.} Proposed by \citet{chen2023accelerating,leviathan2023fast}, speculative sampling independently accepts each node $a_{1:i}$ with probability $\min(1,p(a_i|\con, a_{1:i-1})/q(a_i|\con, a_{1:i-1}))$. Then, it selects the maximal depth node $a_{1:\tau}$ with \textit{all} ancestor nodes accepted. Finally, it samples a correction token from a residual distribution, which is proportional to $\max\{p(\cdot|a_{1:\tau})-q(\cdot|a_{1:\tau}),0\}$ if $\tau<L$ and $p(\cdot|a_{1:\tau})$ otherwise.

\paragraph{Tree verification.} Tree verification \citep{hu2024accelerated} provably improves upon naive speculative sampling by using Monte Carlo tree sampling to increase $\mathbb{E}[\tau]$.

\paragraph{Block verification (BV).} Block verification \citep{sun2024block} provably improves upon naive speculative sampling by relaxing the requirement that all ancestors must be accepted. They recursively define node weights $w(a_{1:i})= \min(1,w(a_{1:i-1})p(a_i|a_{1:i-1})/q(a_i|a_{1:i-1}))$ and independently accept each node $a_{1:i}$ according to these weights. They return the maximal depth node $a_{1:\tau}$ and sample the correction token from a $w$-weighted naive residual.

As \citet{sun2024block} show in their paper, block verification achieves the highest block efficiency among any single-path verification algorithm that only takes in $p$ and $q$ data on $a_{1:L}$, so it is also provably better than tree verification.

\subsection{Multi-Path Algorithms}
\label{subsec:multi-path}

Now, we consider draft trees which are the union of $K$ i.i.d. length $L$ paths $a[k]_{1:L} \sim q_L(\cdot|\con)$. When paths overlap, for every parent-child node pair $(\con_p,\con_c)$ ,the multiplicity of $\con_c$ in the child node list of $\con_p$ is the number of times $\con_c$ appears as a prefix of some $a[k]$. Such a draft tree remains practical for speculative decoding due to GPU parallelization. For moderate $K$ and $L$, there is little overhead in performing the \emph{batched} forward pass across $K$ sequences, relative to a single forward pass \citep{agrawal2024sarathiserve,dao2022flashattention}. 

The majority of multi-path algorithms use optimal transport linear program (OTLP) solvers, which is a specialized next-token predictor tailored to the draft distribution.

\begin{definition}
    An \textbf{OTLP solver} $f_{p,q,k}:\V^k \to \V$ on distributions $p,q \in \Delta(\V)$ with multiplicity $k \in \mathbb{N}$ is a probabilistic function such that $f(X_1,\ldots,X_k)$ follows the $p$ distribution for i.i.d. $X_1,\ldots,X_k\sim q$.
    \label{def:otlp}
\end{definition}

\paragraph{OT-based algorithms.} 
 Any OTLP solver can be used to perform a \textbf{top-down} traversal of the draft tree, starting at the root. Specifically, at each node $\con$, taking in $p(\cdot|\con),q(\cdot|\con)$ and the child node list $\text{ch}(\con)=[(\con,x_i)]_{i=1}^k$, an OT-based solver computes a feasible solution of an optimal transport linear program to probabilistically append a token $t$ to $\con$, progressing to the child node $(\con,t)$ if $t \in \{x_1,\ldots,x_k\}$ and terminating otherwise.  To the best of our knowledge, the unique OT-based algorithms in the i.i.d. multi-path setting are NSS \citep{miao2024specinfer}, SpecInfer \citep{miao2024specinfer}, SpecTr \citep{sun2023spectr}, and Khisti \citep{khisti2025multi}. Also, while naive speculative sampling \citep{leviathan2023fast,chen2023accelerating} was originally presented as a single-path algorithm, it is OT-based and can also be applied to multi-path methods: to distinguish this from Naive, we call it NaiveTree. We review these details in \cref{appendix:otlp-review}.

\paragraph{Traversal verification.} \citet{weng2025traversal} gives the only multi-path algorithm that traverses the draft tree from the \textbf{bottom-up}. When $K=1$, this reduces to block verification.

We stress that these algorithms may outperform each other in different $(p,q)$ regimes, which inherently depend on the model, dataset, and $p$-sampling setting. On the OT-based side, recent work \citep{hu2025towards,thomas2025global} has computed the acceptance rate of an optimal OTLP solver or a near-optimal OTLP solver in certain cases, so a fully optimal OT-based remains infeasible. To the best of our knowledge, there are no existing comparisons between traversal verification and \textit{all} OT-based methods.

\subsection{Other Algorithms}

In the deterministic tree setting, only NSS \citep{miao2024specinfer} works, so it is used in works like EAGLE \citep{li2024eagle}. Some works also mix i.i.d. and deterministic constructions. For example, \citet{hu2025towards} devise an OT-based algorithm when the top-$(k-1)$ draft tokens are chosen deterministically and the last is selected randomly without replacement, with SpecHub \citep{sun2024spechub} having prior considered the case $k=2$. We leave exploration of these methods in the context of dynamic drafting to future work.

\begin{table}[t]
\centering
\caption{Average block efficiency across datasets and sampling configurations of existing verification algorithms. For more detailed data, see Appendix \ref{appendix:full-online-results}.}
\label{tab:block_eff_avg_sorted}
\begin{tabular}{lcccc}
\toprule
\textbf{Method} & \textbf{Qwen} & \textbf{Gemma} & \textbf{Llama} & \textbf{Average} \\
\midrule
NSS & 4.44 & 1.99 & 5.72 & 4.05 \\
BV & 4.24 & 3.30 & 5.37 & 4.30 \\
Khisti & 4.90 & 2.05 & 6.29 & 4.41 \\
NaiveTree & 5.00 & 2.01 & 6.50 & 4.50 \\
Naive & 4.60 & 3.44 & 5.52 & 4.52 \\
SpecInfer & 5.13 & 2.07 & 6.55 & 4.58 \\
SpecTr & 5.11 & 2.05 & 6.67 & 4.61 \\
\textbf{Traversal} & \textbf{5.33} & \textbf{3.81} & \textbf{6.78} & \textbf{5.31} \\
\bottomrule
\end{tabular}
\end{table}



\section{Comparison of Existing Verification Algorithms}
\label{sec:multi-comparison}

In this section, we compare the verification algorithms mentioned in the previous sections. We include two single-path algorithms: original (naive) speculative sampling and block verification (BV); and 6 multi-path algorithms: NSS, NaiveTree, SpecTr, SpecInfer, Khisti, and Traversal.

\subsection{Experimental Setup}
\label{subsec:experimental_setup}

We perform our experiments on three target-draft model pairs: Llama-3 70B/8B-Instruct \citep{grattafiori2024llama}, Gemma-3 27B/270M-IT \citep{gemmateam2025gemma3technicalreport}, and Qwen-2.5 32B/0.5B-Instruct \citep{qwen2025qwen25technicalreport}. Our selection covers three different model families, target model sizes, and different magnitudes of target to draft size ratios ($\sim$9:1, 100:1, and 64:1 respectively). 

We conduct our experiments on 5 datasets: MATH500 \citep{lightman2023lets}, OlympiadBench \citep{he2024olympiadbench}, LiveCodeBench \citep{jain2024livecodebench}, LitBench \citep{fein2025litbenchbenchmarkdatasetreliable}, and Opus \citep{zhang-etal-2020-improving, tiedemann-2012-parallel}. In LiveCodeBench, we use the `code\_generation\_lite' subset; for Opus, we select half of our prompts each from `opus\_books` and `open\_subtitles` with a uniform split of translations between English, French, Spanish, and Italian. Each of these datasets covers a different generative setting. MATH500 and OlympiadBench both cover math, but at differing levels of difficulty; LiveCodeBench is focused on coding; we use the prompts from LitBench to test creative writing; and we use Opus to test performance for translation.

We evaluate methods on 50 prompts per dataset. We test on 8 different sampling configurations: sampling from $M_p$ with temperatures $\{0.2, 0.4, 0.6, 0.8, 1.0, 1.2\}$ and no nucleus sampling, and sampling from $M_p$ with temperature 1.0 and nucleus sampling with $\{0.9, 0.99\}$. We evaluate throughput on machines with two A100-80GB GPUs and an Intel Xeon Gold 6342 CPU (12 cores, 2.80 GHz).

We focus on two important metrics for comparing speculative decoding methods: \textbf{block efficiency} and \textbf{throughput}. Block efficiency is the average number of accepted tokens per speculated block; the greater this number, the fewer target forward passes are required. Block efficiency improves as nodes are added to a tree; however, speedups in practice tend instead to follow a U-curve, as larger trees slow down both drafting and target model forward passes. As such, throughput, measured in tokens per second, is more relevant in deployment. However, its shortcoming is that it is highly sensitive to system-level parameters, such as GPU speed/memory, inference engines, and attention kernels.

\begin{table}[t]
\centering
\caption{Average throughput (tokens per second) across datasets and sampling configurations of existing verification algorithms. For more detailed data, see Appendix \ref{appendix:full-online-results}.}
\label{tab:tps_avg_sorted}
\begin{tabular}{lcccc}
\toprule
\textbf{Method} & \textbf{Qwen} & \textbf{Gemma} & \textbf{Llama} & \textbf{Average} \\
\midrule
NSS & 16.88 & 5.94 & 12.91 & 11.91 \\
Naive & 16.55 & 9.20 & 11.58 & 12.44 \\
Khisti & 18.67 & 5.93 & 13.66 & 12.75 \\
NaiveTree & 19.16 & 6.00 & 14.71 & 13.29 \\
SpecTr & 19.01 & 6.09 & \textbf{14.86} & 13.32 \\
BV & 17.30 & 10.57 & 12.20 & 13.36 \\
SpecInfer & 19.58 & 6.21 & \textbf{14.86} & 13.55 \\
\textbf{Traversal} & \textbf{21.35} & \textbf{11.56} & 14.77 & \textbf{15.89} \\
\bottomrule
\end{tabular}
\end{table}

\subsection{Results}

Our results comparing existing verification algorithms are summarized in \cref{tab:block_eff_avg_sorted} and \cref{tab:tps_avg_sorted}. These tables average across all datasets and sampling configurations per model; the detailed breakdowns are in Appendix \ref{appendix:full-online-results}. In reporting our results, for each set of sampling parameters and dataset, we select the branching factor $K\in[1,4]$ and block length $L\in[0,8]$ that maximizes the block-efficiency or throughput.

We see that Traversal is the best-performing verification algorithm in both block efficiency and throughput, outperforming the second-best method in each by $\sim15\%$. Traversal shows particularly strong performance on Gemma, but is also the top performer for Qwen and only slightly below the best throughput for Llama. Further analysis of full results in Appendix \ref{appendix:full-online-results} shows that Traversal is consistently the best algorithm in both throughput and block efficiency across all sampling configurations tested. We hypothesize this occurs because the top-down approach in OT-based methods makes deep traversal in tree exponentially difficult, whereas the bottom-up approach in Traversal starts at leaf nodes and has a higher chance of accepting longer sequences.

After Traversal, we find that SpecTr, BV, and SpecInfer all perform similarly in average throughput. NSS performs worst, which is expected because it does not use draft probabilities to guide verification.

\section{Delayed Tree Expansion}
\label{sec:delayed-expansion}

While our results in \cref{sec:multi-comparison} show Traversal significantly outperforms OT-based methods in all settings, interestingly, there is no dominant winner among OT-based methods. For throughput, SpecInfer wins for Qwen, Naive wins for Gemma, and SpecTr and SpecInfer tie on Llama, with Khisti not lagging far behind. To explain this phenomenon, we examine acceptance rates of OT-based methods at tree nodes.

The acceptance rate measures how often an OT-based method appends a token that remains on the draft tree. This incorporates the randomness of draft tokens, so it does not actually depend on the observed child nodes, rather only on $p$ and $q$. We show how to compute these rates in \cref{appendix:otlp-accept}.

\begin{definition}
    For any OTLP solver $f_{p,q,k}$, its \textbf{acceptance rate} $\alpha(f_{p,q,k})=\mathbb{P}(f(X_1,\ldots,X_k) \in \{X_1,\ldots,X_k\})$ is the probability that the OTLP solver output token lies among its input tokens, over i.i.d. input samples $X_1,\ldots,X_k \sim q$.
    \label{def:accept}
\end{definition}

\begin{figure*}[ht]
  \centering
  \begin{subfigure}{0.49\textwidth}
    \centering
    \includegraphics[width=\linewidth]{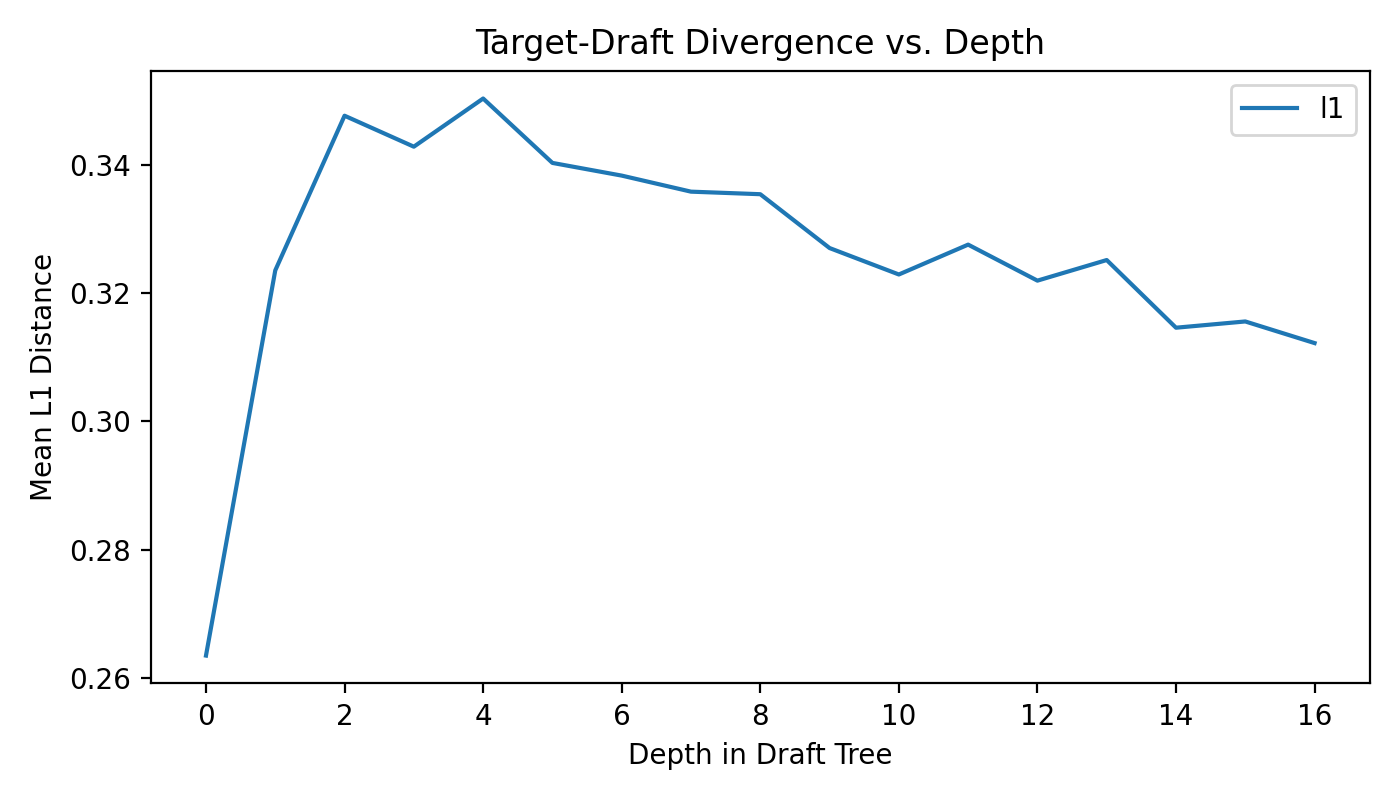}
    \label{fig:nm-mean-l1}
  \end{subfigure}\hfill
  \begin{subfigure}{0.49\textwidth}
    \centering
    \includegraphics[width=\linewidth]{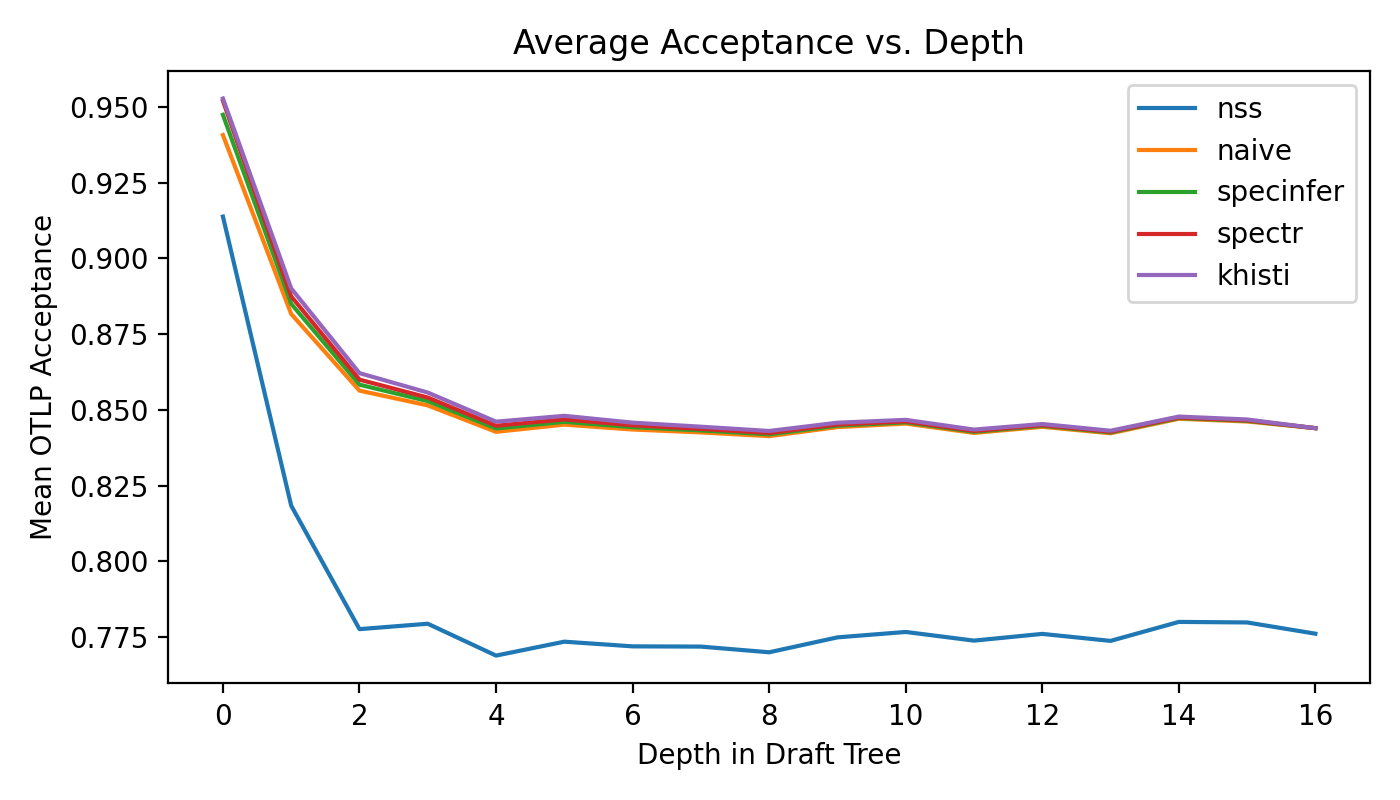}
    \label{fig:nm-mean-accprob}
  \end{subfigure}
  \caption{We generate 200,000+ draft  (Llama-3 8B-Instruct) trees from roots of target model (Llama-3 70B-Instruct) trajectories and compute both L1 target-draft distance and average OTLP acceptances across varying draft tree depths. The divergence between target and draft distributions spikes deeper in the tree, and acceptances across all OTLP methods consistently decrease with depth.}
  \label{fig:nm-side-by-side}
\end{figure*}

We employ the same experimental setup as in \cref{sec:multi-comparison}, but conduct our analysis on \textit{offline} trees generated from over $200,000$ fixed-spaced draft roots along fixed target model trajectories. This allows us to use vLLM \citep{kwon2023efficient} to speed up data generation. Our aggregate results across all $8$ sampling settings for LiveCodeBench, MATH500, and OlympiadBench  on Llama are shown in \cref{fig:nm-side-by-side}, and we also include L1 distance between target and draft distributions. 

We directly observe for all OT-based methods that node-wise acceptance rate rapidly degrade with depth. In other words, while we branch early in the tree often, the acceptance rate improvements later in the tree are often beneficial. This is directly explained by the fact that L1 target-draft deviations increase with depth. For example, the Naive acceptance rate for $k=1$ linearly decreases with L1 distance.

Motivated by these findings, we define a delayed tree, which drafts a partial path and then expands. OTLP solvers can still be used on such a tree while preserving the target distribution, because conditioned on their current node, their output token follows the $p$ distribution.

\begin{definition}
    Fix $K,L_1,L_2 \in \mathbb{N}$ with $K\geq 1$. A \textbf{random $(K,L_1,L_2)$-delayed-tree} at context $\con$ is formed by sampling a length-$L_1$ path $a_{1:L_1} \sim q(\cdot|\con)$ and then branching into $K$ i.i.d. length-$L_2$ paths $a^{(k)}_{1:L_2} \sim q(\cdot|\con,a_{1:L_1})$. Child node lists can have duplicates if paths overlap after the branched node $(\con,a_{1:L_1})$, but not before.
    \label{def:expandtree}
\end{definition}

Now that we have motivated delayed expansion, we discuss its evaluation. How do we estimate the block efficiency of an OT-based method on a delayed tree, when both drafting and verification involve randomness? Motivated by adjacent work in adaptive drafting \citep{chen2024sequoia,brown2024dynamic}, we define the notion of branching probability for an OTLP solver.

\begin{definition}
    For any OTLP solver $f_{p,q,k}$, list of $k$ tokens $\boldsymbol{x}$, and token $t \in \V$, the \textbf{branching probability} to $t$, $B(f_{p,q,k},\boldsymbol{x},t)$, is the probability that $f_{p,q,k}(\boldsymbol{x})=t$.
    \label{def:branch}
\end{definition}

This can be used to compute expected block efficiency by following the law of conditional expectation to separate draft tree randomness and verification randomness: the outer expectation is taken over random $(K,L_1,L_2)$-delayed-trees $\mathcal{T}$, the inner expectation is taken over the randomness present in the OTLP solver, and $p,q$ in $f_{p,q,k}$  represent the implied node distributions $p(\cdot|\con,t_1,\ldots,t_{j-1})$ and $q(\cdot|\con,t_1,\ldots,t_{j-1})$. The last equality holds from the fact that OT-based methods iteratively append the output of $f_{p,q,k}$ to progress down the draft tree.
\begin{align}
    &\mathbb{E}[\tau + 1] = \mathbb{E}_{\mathcal{T}} \left[\mathbb{E}[\tau + 1|\mathcal{T}]\right] =\\
    &\mathbb{E}_{\mathcal{T}} \left[\sum_{\con' \in \mathcal{T}} \mathbb{P}(\text{OTLP solver reaches }\con'|\mathcal{T})\right] =\\
     &\mathbb{E}_{\mathcal{T} } \left[\sum_{\substack{(\con,t_1,\ldots,\\ t_d)\in \mathcal{T}}} \prod_{j=1}^{d} B\left(f_{p,q,k},\text{ch}(\con,t_1,\ldots,t_{j-1}),t_{j}\right)\right]
     \label{eqn:estimator}
\end{align}
In practice, we estimate this quantity by taking a simple average of the inner sum over $s=4$ i.i.d. delayed tree samples $\mathcal{T}_1,\ldots,\mathcal{T}_s$. This estimator does not eliminate drafting variance, but eliminates verification variance and is unbiased.

\section{Neural Delay-and-Branch Predictor}
\label{sec:neural-selector}

Delayed tree expansion, as described in \cref{sec:delayed-expansion}, introduces discrete design choices that can strongly influence end-to-end throughput. For a fixed verification procedure, the throughput of a delayed expansion policy is a tradeoff between two variables: \textit{acceptance}, representing how many tokens in the draft tree are verified, and \textit{latency}, which is the combined cost of drafting and the target pass. Because the divergence between the $p$ and $q$ distributions varies significantly across contexts, a single static configuration of all three parameters $(K,L_1,L_2)$ is not necessarily optimal across all requests. Therefore, we introduce a context-conditioned parameter selector that predicts which delayed expansion parameters to use at each decoding step.

Formally, at each decoding step with root context $\con$, our neural selector chooses the parameters $
a = (K,L_1,L_2) \in \mathcal{A}$, where $\mathcal{A}=\{1,2,3,4\} \times \{0,\ldots,8\}^2$ is the action space representing all supported configurations of the parameters described in \cref{sec:delayed-expansion}. After choosing $a$, we draft the corresponding delayed tree, run a target forward pass, and carry out verification. Ultimately, we choose $a$ to maximize a throughput objective which trades off between block efficiency and latency (\cref{eqn:neural-tps-definition}). As $\mathcal{A}$ is discrete and small, we implement $\pi$ as a categorical policy.

Since the selector must choose $a$ before constructing the draft tree, we restrict it to the following features that are readily available at the root, described in \cref{appendix:neural-details}: \textbf{(i)} draft and target model hidden states at the root and draft states at the previous token, \textbf{(ii)} scalar uncertainty and divergence statistics between $p$ and $q$, such as entropy and KL divergence, \textbf{(iii)} local sampling parameters and context length, and \textbf{(iv)} latency estimates for draft and target forward passes.

We design this policy from a lightweight MLP which independently projects hidden state inputs to a shared dimension $d=128$, applies layer normalization, concatenates standardized scalar features, and applies a two-layer MLP to produce logits over supported actions. To train the selector as described above, we construct a target signal for each action from an offline dataset of speculative decoding traces, taking a root every 16 tokens. For each root $\con$ and action $a$, we store an empirical estimate $\widehat{E}[\tau(\con,a)+1]$ of block efficiency for an $a$-delayed-tree at $\con$ following \cref{eqn:estimator}. To incorporate throughput considerations, we estimate the runtime for each action using latency measurements derived from a microbenchmark "warm-up run", then approximate the total wall-clock time for all necessary forward passes $\hat{T}$ as shown in \cref{eqn:neural-forward-time}.

Given the logits $z_{\theta}(x_i)$, we can then define the policy's per-sample offline throughput estimate at train time as the ratio of expected block efficiency to expected time:

\begin{equation}
\widehat{\text{TPS}}_{\pi}(\con)=\frac
{\sum_{a\in\mathcal{A}}{\pi_{\theta}(a|\con)\widehat{E}[\tau(\con,a)+1]}}
{\sum_{a\in\mathcal{A}}{\pi_{\theta}(a|\con)\widehat{T}(\con,a)}}.
\label{eqn:tps-estimator}
\end{equation}

For each training example, we also have a static baseline configuration $a_i^{base}$ associated with its sampling parameters (temperature and nucleus threshold). Then, we optimize a baseline-aware objective by minimizing

\begin{equation}
-\text{log}\frac{\widehat{\text{TPS}}_{\pi}(\con)}{\widehat{\text{TPS}}_{base}(\con)}.
\end{equation}

In addition to this primary term in the objective, we add the penalty term described in \cref{appendix:neural-details} which encourages the selector to avoid severe regressions below the baseline, even if they occur infrequently.

At inference time, the selector produces $\pi_{\theta}(\cdot \mid \con)$ once per decoding step using the root-level features. We then select the action with the highest assigned probability $\text{argmax}_{a \in \mathcal{A}}\pi_{\theta}(a \mid \con)$, which corresponds to the $(K,L_1,L_2)$ parameters necessary for delayed expansion, before proceeding with the drafting and verification processes.


\subsection{Experiments}
\label{subsec:neural_experiments}

We now evaluate the effectiveness of our neural selector. We follow a similar experimental setup to that described in \cref{subsec:experimental_setup}. We train our parameter selector on offline data generated from 350 prompts per dataset, and measure its performance on a held-out test set.

First, we compare the efficacy of our neural selector (NDE) to the baseline algorithms. The results of this are shown in \cref{tab:block_eff_speedup_iid} and \cref{tab:tps_speedup_iid}. We compute the ratio of improvement in block efficiency and TPS per verification method. We see that NDE improves on the baseline algorithms by around $10\%$ and $25\%$ in block efficiency and throughput on average, across the models tested. Significant improvements are obtained in particular on Gemma. We speculate that this is because our Gemma target/draft model pair has the highest ratio of size difference, and therefore likely has the most divergent proposal distribution, which our neural selector is able to navigate by dynamic parameter selection.

We then compare the best performing existing verification method, Traversal, to our new algorithms. The results of this are shown in \cref{tab:block_eff_avg_nde} and \cref{tab:tps_avg_nde}. We see that the NDE is effective at reducing the gap of the OT methods to Traversal; and now, SpecInfer NDE is able to outperform Traversal in throughput by $\sim5\%$ on average, achieving a new state-of-the-art.

\begin{table}[t]
\centering
\caption{Block efficiency ratio improvement of our NDE (neural dynamic expansion) approach over the baseline algorithms.}
\label{tab:block_eff_speedup_iid}
\begin{tabular}{lcccc}
\toprule
\textbf{Method} & \textbf{Qwen} & \textbf{Gemma} & \textbf{Llama} & \textbf{Average} \\
\midrule
Khisti NDE & 1.01 & 1.90 & 1.09 & 1.19 \\
NaiveTree NDE & 1.07 & 0.99 & 1.17 & 1.09 \\
NSS NDE & 0.97 & 1.68 & 1.01 & 1.11 \\
SpecInfer NDE & 0.98 & 1.65 & 1.01 & 1.10 \\
SpecTr NDE & 0.97 & 1.68 & 1.01 & 1.10 \\
\bottomrule
\end{tabular}
\end{table}

\begin{table}[t]
\centering
\caption{Throughput (tokens per second) ratio improvement of our NDE (neural dynamic expansion) approach over the baseline algorithms.}
\label{tab:tps_speedup_iid}
\begin{tabular}{lcccc}
\toprule
\textbf{Method} & \textbf{Qwen} & \textbf{Gemma} & \textbf{Llama} & \textbf{Average} \\
\midrule
Khisti NDE & 0.99 & 2.17 & 0.97 & 1.16 \\
NaiveTree NDE & 1.31 & 1.40 & 1.21 & 1.30 \\
NSS NDE & 1.20 & 2.09 & 0.96 & 1.26 \\
SpecInfer NDE & 1.15 & 2.14 & 0.96 & 1.23 \\
SpecTr NDE & 1.14 & 2.17 & 0.96 & 1.23 \\
\bottomrule
\end{tabular}
\end{table}

\begin{table}[t]
\centering
\caption{Average block efficiency across datasets and sampling configurations. Traversal is best existing algorithm; NDE (neural delayed expansion) is ours. For more detailed data, see Appendix \ref{appendix:full-online-results}.}
\label{tab:block_eff_avg_nde}
\begin{tabular}{lcccc}
\toprule
\textbf{Method} & \textbf{Qwen} & \textbf{Gemma} & \textbf{Llama} & \textbf{Average} \\
\midrule
Traversal & \textbf{5.33} & 3.81 & 6.78 & \textbf{5.31} \\
\midrule
NSS NDE & 4.31 & 3.34 & 5.78 & 4.48 \\
NaiveTree NDE & 4.94 & 3.42 & 6.47 & 4.94 \\
SpecInfer NDE & 5.03 & 3.42 & 6.64 & 5.03 \\
SpecTr NDE & 4.97 & 3.44 & 6.74 & 5.05 \\
Khisti NDE & 4.97 & \textbf{3.89} & \textbf{6.84} & 5.23 \\
\bottomrule
\end{tabular}
\end{table}

\begin{table}[t]
\centering
\caption{Average throughput (tokens per second) across datasets and sampling configurations. Traversal is best existing algorithm; NDE (neural delayed expansion) is ours. For more detailed data, see Appendix \ref{appendix:full-online-results}.}
\label{tab:tps_avg_nde}
\begin{tabular}{lcccc}
\toprule
\textbf{Method} & \textbf{Qwen} & \textbf{Gemma} & \textbf{Llama} & \textbf{Average} \\
\midrule
Traversal & 21.35 & 11.56 & \textbf{14.77} & 15.89 \\
\midrule
Khisti NDE & 18.41 & 12.84 & 13.24 & 14.83 \\
NSS NDE & 20.29 & 12.40 & 12.45 & 15.05 \\
NaiveTree NDE & 21.67 & 12.86 & 14.01 & 16.18 \\
SpecTr NDE & 21.74 & 13.22 & 14.23 & 16.40 \\
SpecInfer NDE & \textbf{22.54} & \textbf{13.26} & 14.27 & \textbf{16.69} \\
\bottomrule
\end{tabular}
\end{table}

\section{Conclusion}

We conducted the first systematic analysis of all i.i.d. verification algorithms in multi-path speculative decoding. We discovered that Traversal Verification significantly outperforms all other methods in our evaluation, which are OT-based. By analyzing target-draft divergence and tree node acceptance rates, we found this was caused by aggressive branching early in the draft tree, which wastes compute on minimal acceptance gains. To overcome this shortcoming, we introduce the idea of dynamic delayed draft tree expansion, and trained a neural selector on each OT-based method to optimally select when and how to expand i.i.d. rollouts in a tree based on a corresponding acceptance length objective. With this neural selector, we demonstrated that OT-based methods can outperform Traversal Verification. Future work could explore a similar neural selector for Traversal Verification, or consider more robust training objectives for expected acceptance length. 

\section{Impact Statement}

This paper presents work whose goal is to advance the field
of Machine Learning. There are many potential societal
consequences of our work, none which we feel must be
specifically highlighted here.

\FloatBarrier

\bibliography{example_paper}
\bibliographystyle{icml2026}

\newpage
\appendix
\onecolumn

\section{Orthogonal Drafting Work}
\label{appendix:orthogonal-work}

Various avenues of work improve drafting or its integration with other phases without altering the tree construction. Some even fundamentally alter the draft-verify regime. We emphasize that many of these works are orthogonal to ours, and can be combined with dynamic delayed drafting to further improve throughput. We leave exploration of these ideas to future work.

\paragraph{Improving drafting without tree control.} EAGLE \citep{li2024eagle} trains a lightweight draft model on features to propose many high-quality draft tokens per step. EAGLE-3 \citep{li2025eagle3} scales EAGLE with test-time compute to improve draft quality under fixed latency. Kangaroo \citep{liu2024kangaroo} employs self-speculation to train an adaptive layer on a subnetwork of the target model, to use as a draft model. Cacheback \citep{ma2025cacheback} uses an LRU cache of n-grams to accelerate drafting. Cascade drafting \citep{chen2024cascade} uses horizontal and vertical cascading, with larger drafts for crucial earlier tokens and smaller drafts for later tokens. Ouroboros \citep{zhao2024ouroboros} drafts and verifies phrases rather than tokens. Previous work also considers retrieval \citep{he2023rest}, hierarchical drafting \citep{sun2024triforce}, and layer skipping \citep{zhang2023draft,elhoushi2024layerskip}. Our work is agnostic to how draft candidates are produced, so we do not compare against these methods.

\paragraph{Pipelining and parallelism.} Some hardware-aware work focuses on reducing rollbacks and pipeline stalls. Pearl \citep{liu2024pearl} reduces mutual waiting between drafting and target passes by overlapping computation with pre/post-verification pipelining, and tunes adaptive draft lengths for end-to-end latency instead of block efficiency. SpecBranch \citep{shen2025speculative} improves Pearl by introducing branch-parallel speculative branches with rollback-aware hybrid drafts, explicitly trading off between parallelism and rollback costs. These are orthogonal to and can be integrated into our work, since we optimize for drafting and these optimize for scheduling once drafting is fixed.

\paragraph{Alternative multi-token prediction.} Speculative decoding is not the only method that decodes multiple tokens per target call. Parallel decoding \citep{stern2018blockwise} proposes many tokens and uses parallel scoring to reduce autoregressive dependence, but does not faithfully maintain the target distribution. Medusa \citep{cai2024medusa} trains multiple independent heads over the draft model to propose draft tokens, and Hydra \citep{ankner2024hydra} extends on Medusa with sequential decoding. Multi-token prediction \citep{gloeckle2024better,samragh2025your} trains auxiliary heads without a draft model to predict multiple future tokens from a single forward pass. Lookahead decoding \citep{fu2024break} generates many disjoint n-grams in one parallelized step, and performs verification to maintain the target distribution.

\newpage

\section{Review of OT-Based Methods}
\label{appendix:otlp-review}

Here, we review the five OT-based verification algorithms. Each involves traversing the draft tree top-down based on its corresponding OTLP solver, as described in \cref{subsec:multi-path}. Below, $p$ and $q$ denote the target and draft distributions, and we use the shorthand notation $\boldsymbol{x}_+ = \max\{\boldsymbol{x},0\}$ for vectors $\boldsymbol{x}$. We also use $\propto$ to denote normalizing a distribution.

\subsection{NSS}

NSS\footnote{\citet{miao2024specinfer} called this "Naive Speculative Sampling", but we refer to standard speculative sampling as naive, and call this NSS.} is the simplest OT-based algorithm, explicitly defined in \citep{miao2024specinfer}. Ignoring the draft tokens, NSS always directly samples a token from the target distribution, as in \cref{alg:otlp-nss}. Because NSS does not depend at all on draft model probabilities, it can be used in any drafting regime. Thus, methods like EAGLE-2, which use deterministic trees, use NSS.

\begin{algorithm}[ht]
\caption{NSS OTLP Solver}
\label{alg:otlp-nss}
\begin{algorithmic}[1]
\INPUT Target distribution $p\in\Delta(\V)$, draft distribution $q\in\Delta(\V)$, i.i.d. draft tokens $X_1,\ldots,X_k \sim q$
\OUTPUT Token $Y\in\V$
\STATE Sample $Y \sim p$
\STATE \textbf{return} $Y$
\end{algorithmic}
\end{algorithm}

\subsection{Naive}

Naive speculative sampling, as we described in \cref{subsec:single-path}, was originally formulated as a single-path method \citep{leviathan2023fast,chen2023accelerating}. However, one can observe that independently accepting draft tree nodes, and then selecting the maximum depth node with all accepted ancestors, is equivalent to performing a top-down OT-based draft tree traversal when the draft tree is a single path. The output of the OTLP solver here is given by either accepting the single child token, or sampling from a residual. This can be extended to a multi-path method in \cref{alg:otlp-naive} by following the same OTLP solver only on the first draft token $X_1$, but still allowing one to output and traverse to other draft tokens $X_2,\ldots,X_k$ when sampling from the residual. Note that while the output of NSS is independent of all draft tokens, the output of Naive is independent of all but the first draft token.

\begin{algorithm}[ht]
\caption{Naive OTLP Solver (Speculative Decoding)}
\label{alg:otlp-naive}
\begin{algorithmic}[1]
\INPUT Target distribution $p\in\Delta(\V)$, draft distribution $q\in\Delta(\V)$, i.i.d. draft tokens $X_1,\ldots,X_k \sim q$
\OUTPUT Token $Y\in\V$
\STATE Sample $U \sim \mathcal{U}([0,1])$
\IF{$U \leq p(X_1) / q(X_1)$}
    \STATE \textbf{return} $X_1$
\ENDIF
\STATE $p_{\text{res}} \propto (p-q)_+$
\STATE Sample $Y \sim p_{\text{res}}$
\STATE \textbf{return} $Y$
\end{algorithmic}
\end{algorithm}

\subsection{SpecTr}

SpecTr uses the OTLP solver K-SEQ, described in detail by \citet{sun2023spectr}, so we recap the sampling procedure below and forgo proofs of losslessness. First, we compute the division factor $\rho^\star$. For each $\rho$, define the quantities
\begin{align}
    \beta(\rho) &= \sum_{x \in \V} \min\left(\rho^{-1}p(x), q(x)\right), \\
    p_{\text{acc}}(\rho) &= 1-(1-\beta(\rho))^k.
\end{align}The division factor is the root of $\rho \mapsto p_{\text{acc}}(\rho)-\rho \beta(\rho)$, which is monotone decreasing on $[1,k]$, so binary search can be used to efficiently compute $\rho^\star$. Then, a $\rho^\star$-weighted version of the Naive OTLP solver is performed for multiple rounds, as in \cref{alg:otlp-spectr}. This OTLP solver precisely reduces to the Naive solver when $k=1$. Note that $\rho^\star$ up to $k$ can be increased and this will remain an OTLP solver, but the optimal acceptance rate (see \cref{appendix:otlp-accept}) is provably achieved at $\rho^\star$.

\begin{algorithm}[ht]
\caption{SpecTr OTLP Solver (K-SEQ)}
\label{alg:otlp-spectr}
\begin{algorithmic}[1]
\INPUT Target distribution $p\in\Delta(\V)$, draft distribution $q\in\Delta(\V)$, draft tokens $X_1,\dots,X_k$
\OUTPUT Token $Y\in\V$
\STATE Solve $p_{\mathrm{acc}}(\rho)=\rho \beta(\rho)$ for $\rho^\star$ by binary search
\STATE $\beta \gets \sum_{t\in\V} \min\left(p(t)/\rho^\star,\,q(t)\right)$
\STATE $p_{\mathrm{acc}} \gets 1-(1-\beta)^k$
\STATE $\gamma \gets p_{\mathrm{acc}} / \beta$
\FOR{$i=1$ to $k$}
    \STATE Sample $U_i \sim \mathcal{U}([0,1])$
    \IF{$ \rho^\star U_i  \le p(X_i) / q(X_i)$}
        \STATE \textbf{return} $X_i$
    \ENDIF
\ENDFOR
\STATE $p_{\text{res}} \propto (p - \min(p/\rho^\star,q)\cdot\gamma)_+$
\STATE Sample $Y \sim p_{\text{res}}$
\STATE \textbf{return} $Y$
\end{algorithmic}
\end{algorithm}

\subsection{SpecInfer}

SpecInfer is described alongside NSS in \citet{miao2024specinfer}. Similar to SpecTr, this algorithm reduces to Naive when $k=1$. While both SpecTr and SpecInfer involve at most $k$ rounds of potential acceptances before sampling from the residual, SpecTr only computes the residual at the end, whereas SpecInfer updates it at each round via uniform child selection.

\begin{algorithm}[ht]
\caption{SpecInfer OTLP Solver}
\label{alg:otlp-specinfer}
\begin{algorithmic}[1]
\INPUT Target distribution $p\in\Delta(\V)$, draft distribution $q\in\Delta(\V)$, draft tokens $X_1,\dots,X_k$
\OUTPUT Token $Y\in\V$
\STATE $S \gets [X_1,\dots,X_k]$
\WHILE{$S \neq [\ ]$}
    \STATE Sample $x \sim \mathcal{U}(S)$ and $U \sim \mathcal{U}([0,1])$
    \IF{$U \le p(x) / q(x)$}
        \STATE \textbf{return} $x$
    \ENDIF
    \STATE $p \propto (p-q)_+$
    \STATE Remove one occurrence of $x$ from $S$
\ENDWHILE
\STATE Sample $Y \sim p$
\STATE \textbf{return} $Y$
\end{algorithmic}
\end{algorithm}

\subsection{Khisti}

Khisti first solves a truncated OTLP based on $p,q,k$ to generate an importance-weighted distribution $r$. Then, they sample a token $x \sim r$ by using pairwise tournament selection on $X_{1:k}$, and perform naive speculative sampling, with $r$ replacing the draft $q$ and $[x]$ replacing the draft tokens $[X_1,\ldots,X_k]$. For further details, see Section 4 of \citet{khisti2025multi}.

\begin{algorithm}[ht]
\caption{Khisti OTLP Solver}
\label{alg:otlp-khisti}
\begin{algorithmic}[1]
\INPUT Target distribution $p\in\Delta(\V)$, draft distribution $q\in\Delta(\V)$, draft tokens $X_1,\dots,X_k$
\OUTPUT Token $Y\in\V$
\STATE $r \gets \textsc{Khisti\_Importance\_Sample}(p,q,k)$
\STATE $x \gets \textsc{Khisti\_Tournament\_Select}(r,X_{1:k})$
\STATE \textbf{return} \textsc{Naive\_OTLP\_Solver}$(p,r,[x])$
\end{algorithmic}
\end{algorithm}

\newpage

\section{OTLP Acceptances}
\label{appendix:otlp-accept}

Now, for each of the five OT-based verification algorithms from \cref{appendix:otlp-review}, we show how to compute the acceptance rate as defined in \cref{def:accept}. We use the exact same notation. These algorithms can all be obtained by following the logic of the OTLP solver and adding the probabilities of accepting tokens at various rounds and sampling draft tokens from the residual. We have empirically confirmed their accuracy with Monte Carlo sampling. 

\textbf{There is one caveat: one cannot get an exact acceptance rate for Khisti efficiently.} However, it is possible to obtain a lower bound by simply ignoring the residual contribution for draft tokens other than $x$.

\begin{algorithm}[H]
\caption{NSS Acceptance}
\label{alg:accept-nss}
\begin{algorithmic}[1]
\INPUT  Target distribution $p\in\Delta(\V)$, draft distribution $q\in\Delta(\V)$, $k\geq 1$
\STATE \textbf{return}  $\sum_{t\in\V} p(t)\big(1-(1-q(t))^k\big)$
\end{algorithmic}
\end{algorithm}

\begin{algorithm}[H]
\caption{Naive Acceptance}
\label{alg:accept-naive}
\begin{algorithmic}[1]
\INPUT Target distribution $p\in\Delta(\V)$, draft distribution $q\in\Delta(\V)$, $k\geq 1$
\STATE \textbf{return} $\sum_{t\in\V}\min(p(t),q(t)) + \sum_{t\in\V}(p(t)-q(t))_+\big(1-(1-q(t))^{k-1}\big)$ 
\end{algorithmic}
\end{algorithm}

\begin{algorithm}[H]
\caption{SpecTr Acceptance}
\label{alg:accept-spectr}
\begin{algorithmic}[1]
\INPUT Target distribution $p\in\Delta(\V)$, draft distribution $q\in\Delta(\V)$, $k\geq 1$
\STATE Solve $p_{\mathrm{acc}}(\rho)=\rho \beta(\rho)$ for $\rho^\star$ by binary search
\STATE $\beta \gets \sum_{t\in\V} \min\left(p(t)/\rho^\star,\,q(t)\right)$
\STATE $p_{\mathrm{acc}} \gets 1-(1-\beta)^k$
\STATE $\gamma \gets p_{\mathrm{acc}} / \beta$
\STATE $p_{\text{res}} \propto (p - \min(p/\rho^\star,q)\cdot\gamma)_+$
\STATE $r\gets (q-p/\rho^\star)_+/(1-\beta)$
\STATE \textbf{return} $p_{\mathrm{acc}} + (1-p_{\mathrm{acc}})\sum_{t\in\V}p_{\text{res}}(t)\big(1-(1-r(t))^k\big)$
\end{algorithmic}
\end{algorithm}

\begin{algorithm}[H]
\caption{SpecInfer Acceptance}
\label{alg:accept-specinfer}
\begin{algorithmic}[1]
\INPUT Target distribution $p\in\Delta(\V)$, draft distribution $q\in\Delta(\V)$, $k\geq 1$
\STATE $p_{\mathrm{rej}} \gets 1, m \gets \mathbf{1}$
\FOR{$i=1$ to $k$}
    \STATE $r \gets \sum_{t\in\V}\min(p(t),q(t))$
    \STATE $p_{\mathrm{rej}} \gets p_{\mathrm{rej}}\cdot(1-r)$
    \STATE $m \gets m \odot (\mathbf{1}-(q-p)_+/(1-r))$
    \STATE $p \propto (p-q)_+$
\ENDFOR
\STATE $\alpha \gets (1-p_{\mathrm{rej}}) + p_{\mathrm{rej}} \sum_{t\in\V}p(t)\big(1-m(t)\big)$
\STATE \textbf{return} $\alpha$
\end{algorithmic}
\end{algorithm}

\begin{algorithm}[H]
\caption{Khisti Acceptance Lower Bound}
\label{alg:accept-khisti}
\begin{algorithmic}[1]
\INPUT Target distribution $p\in\Delta(\V)$, draft distribution $q\in\Delta(\V)$, $k\ge 1$
\STATE $r \gets \textsc{Khisti\_Importance\_Sample}(p,q,k)$
\STATE \textbf{return} $\sum_{t\in\V}\min(p(t),r(t))$
\end{algorithmic}
\end{algorithm}

\FloatBarrier
\newpage

\section{OTLP Branching}
\label{appendix:otlp-branch}

Next, for the five OT-based verification algorithms from \cref{appendix:otlp-review}, we compute branching probabilities from \cref{def:branch}. We again use the notation from \cref{appendix:otlp-review}. These algorithms can similarly be obtained by following the logic of the OTLP solver and mixing the branching probabilities from round-wise token acceptances with the final residual. We have confirmed these algorithms are accurate with Monte Carlo sampling. Unlike acceptance computation, Khisti does have an exact algorithm, by extending tournament selection to a distribution on the output of tournament selection conditioned on the draft tokens.

\begin{algorithm}[H]
\caption{NSS Branching}
\label{alg:branch-nss}
\begin{algorithmic}[1]
\INPUT Target distribution $p\in\Delta(\V)$, draft distribution $q\in\Delta(\V)$, draft tokens $X_1,\dots,X_k$
\STATE \textbf{return} $\{X_i \mapsto p(X_i)\}$
\end{algorithmic}
\end{algorithm}

\begin{algorithm}[H]
\caption{Naive Branching}
\label{alg:branch-naive}
\begin{algorithmic}[1]
\INPUT Target distribution $p\in\Delta(\V)$, draft distribution $q\in\Delta(\V)$, draft tokens $X_1,\dots,X_k$
\STATE $a \gets \min(1,p(X_1)/q(X_1))$
\STATE $p_{\text{res}} \propto (p-q)_+$
\STATE \textbf{return} $\{X_i \mapsto (1-a)p_{\text{res}}(X_i) + a\cdot\mathbf{1}\{X_i=X_1\}\}$
\end{algorithmic}
\end{algorithm}

\begin{algorithm}[H]
\caption{SpecTr Branching}
\label{alg:branch-spectr}
\begin{algorithmic}[1]
\INPUT Target distribution $p\in\Delta(\V)$, draft distribution $q\in\Delta(\V)$, draft tokens $X_1,\dots,X_k$
\STATE Solve $p_{\mathrm{acc}}(\rho)=\rho \beta(\rho)$ for $\rho^\star$ by binary search
\STATE $\beta \gets \sum_{t\in\V} \min\left(p(t)/\rho^\star,\,q(t)\right)$
\STATE $p_{\mathrm{acc}} \gets 1-(1-\beta)^k$
\STATE $\gamma \gets p_{\mathrm{acc}} / \beta$
\STATE $p_{\text{res}} \propto (p - \min(p/\rho^\star,q)\cdot\gamma)_+$
\FOR{$i=1$ to $k$}
\STATE $a_i \gets \min(1,p(X_i)/(\rho^\star q(X_i)))$
\ENDFOR
\STATE \textbf{return} $\{X_i \mapsto \sum_{j=1}^k a_j \mathbf{1}\{X_j=X_i\}\prod_{l<j}(1-a_l) +p_{\text{res}}(X_i)\prod_{l=1}^k(1-a_l)\}$
\end{algorithmic}
\end{algorithm}

\begin{algorithm}[H]
\caption{SpecInfer Branching}
\label{alg:branch-specinfer}
\begin{algorithmic}[1]
\INPUT Target distribution $p\in\Delta(\V)$, draft distribution $q\in\Delta(\V)$, draft tokens $X_1,\dots,X_k$
\STATE $p_0 \gets p$ 
\FOR{$i=1$ to $k$}
\STATE $p_i \propto (p_{i-1}-q)_+$
\STATE $a_i \gets \min(\boldsymbol{1},p_{i-1}/q)$
\ENDFOR
\FOR{$i=k$ to $0$}
\FOR{multisets  $S \subseteq [X_1,\ldots,X_k]$ s.t. $|S| = k-i$}
\FOR{$x \in \{X_1,\ldots,X_k\}$}

\IF{$i=k$}
\STATE $\mathcal{B}_i(S;x)\gets p_k(x)$
\ELSE
\STATE $\mathcal{B}_i(S;x)\gets |S|^{-1}\sum_{t\in S}(a_i(t)\mathbf{1}\{t=x\}+(1-a_i(t))\mathcal{B}_{i+1}(S\setminus\{t\};x))$
\ENDIF
\ENDFOR
\ENDFOR
\ENDFOR
\STATE \textbf{return} $\{X_i \mapsto \mathcal{B}_0([X_1,\dots,X_k];X_i)\}$
\end{algorithmic}
\end{algorithm}

\begin{algorithm}[H]
\caption{Khisti Branching}
\label{alg:branch-khisti}
\begin{algorithmic}[1]
\INPUT Target distribution $p\in\Delta(\V)$, draft distribution $q\in\Delta(\V)$, draft tokens $X_1,\dots,X_k$
\STATE $r \gets \textsc{Khisti\_Importance\_Sample}(p,q,k)$
\FOR{$x \in \{X_1,\ldots,X_k\}$}
\STATE $\pi_x \gets \mathbb{P}(\textsc{Khisti\_Tournament\_Select}(r,X_{1:k})=x)$
\ENDFOR
\STATE \textbf{return} $\sum_{x\in\{X_1,\dots,X_k\}}\pi_x \cdot \textsc{Naive\_Branching}(p,r,[x])$

\end{algorithmic}
\end{algorithm}

\newpage 

\section{Neural Selector Details}
\label{appendix:neural-details}

Let $\mathcal{A}$ be the space of all supported actions for delayed expansion, which we define as the Cartesian product
\begin{equation}
A = \{1,...,K_{\text{max}}\} \times \{0,...,L_{\text{1,max}}\} \times \{0,...,L_{\text{2,max}}\},
\end{equation}

where an action $a=(K,L_1,L_2) \in \mathcal{A}$ specifies the branching factor $K$, the trunk draft length $L_1$ and the branch draft length $L_2$. When using delayed tree expansion, we aim to maximize the throughput as measured in tokens per second (TPS):

\begin{equation}
\max_{\pi}\mathbb{E}_{\con}\!\left[
\frac{\mathbb{E}[\tau+1\mid \con,a]}
     {\mathbb{E}[T(\con,a)]}
\right],
\qquad a \sim \pi(\cdot \mid \con),
\label{eqn:neural-tps-definition}
\end{equation}

where $\mathbb{E}[\tau+1\mid \con,a]$ is the block efficiency at the context $\con$ having chosen the drafting action $a$, and $\mathbb{E}[T(\con,a)]$ represents the expected wall-time (in seconds) of the necessary draft and target passes performed at context $\con$ under the drafting action $a$. The outer expectation is taken over the distributions of contexts $\con$ we would observe across all inputs.

To this end, we define a selector policy which takes in the following features as input:
\begin{itemize}
    \item \textbf{Hidden-state features} $h^{p}_{\mathrm{prev}}$, $h^{q}_{\mathrm{prev}}$ of the target and draft model at the preceding token, and $h^{q}_{\mathrm{cur}}$, the draft model features at the root token\footnote{We cannot use target features here because the root token is missing a KV cache row from the last iteration, which would require an extra target forward pass. This is also the case for drafting, but an extra draft forward pass is relatively cheap.}, all of which are later projected to a shared dimension $d$,
    \item \textbf{Scalar uncertainty features}, which include the entropies $H(p_{prev}), H(q_{prev}), H(q_{root})$, local divergences $KL(p_{prev}\|q_{prev}), KL(q_{prev}\|p_{prev})$, and a local $L^1$ distance $\|p_{prev}-q_{prev}\|$,
    \item \textbf{Local parameter features}, i.e. the context length $|x|$ and the sampling hyperparameters -- temperature $T$ and nucleus threshold $p_{top}$, and
    \item \textbf{Latency estimates} for a draft and target forward pass at the current context length, to capture hardware-dependent effects.
\end{itemize}
 
We design this selector as an MLP policy with separate projections for each hidden-state feature block. Let $\phi_1, \phi_2, \phi_3$ be the linear projections from each representation vector to a shared dimension $d=128$, each followed by layer normalization (LN), and $s(x)$ be the (standardized) scalar feature vector. The policy then computes

\begin{equation}
z_{\theta}(x)
= MLP\left(
    \big[
        LN(\phi_1(h^p_{\text{prev}}));
        LN(\phi_2(h^q_{\text{prev}})); LN(\phi_3(h^q_{\text{cur}}));
        s(x)
    \big]
\right),
\end{equation}

where the MLP has two hidden layers (of sizes 512 and 32 respectively), with GELU activations \cite{hendrycks2023gaussianerrorlinearunits} and dropout. The output layer of this model is a vector of $|A|$ logits, representing the probabilities of each action in $A$ being optimal with respect to throughput. Using this structure, the overhead for the selector forward pass is negligible compared to any draft or target model pass.

Let $t_p(l)$ be the measured target model time in seconds for a forward pass at context length $l$, and let $t_q(l)$ be the draft model time. Then, for an action $a=(K,L_1,L_2)$ at context length $l=|x|$ we approximate total wall-clock time as

\begin{equation}
\hat{T}(x,a) = \underbrace{\sum_{j=0}^{L_1-1} t_q(l+j)}_{\text{trunk drafting}} + \underbrace{\sum_{j=0}^{L_2-1} t_q(l+L_1+j*K)}_{\text{branch drafting}}  + \underbrace{t_p(l+L_1+K L_2)}_{\text{target forward pass}}.
\label{eqn:neural-forward-time}
\end{equation}

\newpage

which we use to estimate total throughput as shown in \cref{eqn:tps-estimator}. Given this estimate $\widehat{TPS}_{\pi}(\con)$ and the throughput of the baseline action $\widehat{TPS}_{base}(\con)$, we ultimately define the policy's objective as minimizing:

\begin{equation}
\mathcal{L}
=
\frac{1}{B}\sum_{i=1}^{B}
\left(
-\log \frac{\widehat{TPS}_{\pi}(x_i)}{\widehat{TPS}_{base}(x_i)}
\right)
\;+\;
\lambda \cdot
\frac{1}{|\mathcal{I}_\alpha|}
\sum_{i \in \mathcal{I}_\alpha}
\left(
\max\left\{
1 - \frac{\widehat{TPS}_{\pi}(x_i)}{\widehat{TPS}_{base}(x_i)},
0
\right\}
\right)^2,
\label{eqn:neural-full-loss}
\end{equation}

where $B$ is the minibatch size, $\lambda$ is the penalty weight for regressing below the baseline, and $\mathcal{I}_\alpha$ are the indices of the largest $\alpha$ fraction of the penalty terms $(\max\left\{1 -\frac{\widehat{TPS}_{\pi}(x_i)}{\widehat{TPS}_{base}(x_i)},0\right\})^2$ within the minibatch, which penalize worst-case throughput regressions (relative to the baseline) to prevent the policy from trading occasional but large throughput regressions for marginal average improvements. 

\newpage
\FloatBarrier

\section{Extended Online Experimental Results}
\label{appendix:full-online-results}

\begin{table}[!htbp]
\tiny
\setlength{\tabcolsep}{3pt}
\begin{tabular}{lrrrrr rrrrr rrrrr}
\toprule
 & \multicolumn{5}{c}{Qwen-2.5 (32B / 500M)}
 & \multicolumn{5}{c}{Gemma-3 (27B / 270M)}
 & \multicolumn{5}{c}{Llama-3 (70B / 8B)} \\
\cmidrule(lr){2-6} \cmidrule(lr){7-11} \cmidrule(lr){12-16}
 & Writing & Coding & Translation & Math (E) & Math (H)
 & Writing & Coding & Translation & Math (E) & Math (H)
 & Writing & Coding & Translation & Math (E) & Math (H) \\
\midrule
Khisti, delayed expansion
 & 14.21 & 17.81 & 14.16 & 23.05 & 22.80
 & 8.93 & 9.90 & 10.73 & 17.39 & 17.28
 & 11.00 & 14.24 & 13.93 & 13.75 & 13.26 \\
Khisti
 & 11.22 & 19.87 & 13.88 & 24.29 & 24.07
 & 3.53 & 5.08 & 8.88 & 6.60 & 5.58
 & 11.71 & 14.25 & 14.23 & 14.18 & 13.92 \\
\midrule
NaiveTree, delayed expansion
 & 14.57 & 21.91 & 16.29 & 28.89 & 26.69
 & 9.95 & 8.45 & 11.95 & 15.84 & 18.10
 & 11.31 & 15.66 & 15.04 & 14.76 & 13.29 \\
NaiveTree
 & 10.25 & 19.08 & 13.28 & 27.52 & 25.67
 & 3.48 & 5.26 & 9.74 & 6.48 & 5.04
 & 11.77 & 16.10 & 15.71 & 15.68 & 14.28 \\
Naive
 & 7.90 & 17.47 & 9.76 & 24.84 & 22.77
 & 4.72 & 6.79 & 7.30 & 13.15 & 14.07
 & 8.65 & 13.05 & 12.72 & 12.42 & 11.04 \\
\midrule
NSS, delayed expansion
 & 11.85 & 20.61 & 16.93 & 26.96 & 25.10
 & 10.17 & 6.39 & 11.89 & 15.82 & 17.76
 & 9.77 & 14.75 & 14.39 & 12.25 & 11.10 \\
NSS
 & 8.00 & 16.28 & 12.67 & 24.68 & 22.76
 & 3.50 & 5.34 & 9.87 & 6.13 & 4.85
 & 10.08 & 15.36 & 15.03 & 12.86 & 11.23 \\
\midrule
SpecInfer, delayed expansion
 & 15.52 & 23.16 & 17.39 & 29.31 & 27.33
 & 10.02 & 9.85 & 12.10 & 16.53 & 17.77
 & 11.80 & 15.77 & 15.08 & 14.79 & 13.90 \\
SpecInfer
 & 10.77 & 19.76 & 13.43 & 27.81 & 26.14
 & 3.59 & 5.34 & 9.92 & 6.55 & 5.67
 & 12.29 & 16.43 & 15.77 & 15.52 & 14.27 \\
\midrule
SpecTr, delayed expansion
 & 15.58 & 22.03 & 17.26 & 27.33 & 26.50
 & 10.00 & 9.87 & 12.03 & 16.24 & 17.97
 & 11.73 & 15.59 & 14.85 & 15.04 & 13.92 \\
SpecTr
 & 10.94 & 19.31 & 13.29 & 26.56 & 24.96
 & 3.54 & 5.38 & 9.80 & 6.44 & 5.26
 & 12.21 & 16.23 & 15.63 & 15.70 & 14.55 \\
\midrule
BV
 & 9.53 & 18.24 & 11.22 & 24.69 & 22.81
 & 5.82 & 7.66 & 8.57 & 15.58 & 15.25
 & 9.57 & 13.38 & 12.94 & 13.07 & 12.04 \\
\midrule
Traversal, $K$=2
 & 11.11 & 20.47 & 12.94 & 27.44 & 25.56
 & 6.31 & 8.01 & 9.07 & 14.24 & 16.08
 & 10.90 & 14.81 & 14.17 & 14.74 & 13.84 \\
Traversal, $K$=3
 & 11.47 & 21.34 & 14.23 & 29.05 & 26.63
 & 6.51 & 8.55 & 9.46 & 15.09 & 17.21
 & 11.95 & 15.39 & 14.66 & 14.55 & 13.86 \\
Traversal, $K$=4
 & 11.95 & 22.90 & 14.97 & 29.41 & 27.49
 & 6.56 & 8.94 & 9.73 & 15.35 & 17.23
 & 12.35 & 15.74 & 15.19 & 15.39 & 15.18 \\
\bottomrule
\end{tabular}
\caption{Tokens per second by dataset and algorithm}
\label{tab:tps_by_dataset_and_algorithm}
\end{table}

\begin{table}[!htbp]
\tiny
\setlength{\tabcolsep}{3pt}
\begin{tabular}{lrrrrr rrrrr rrrrr}
\toprule
 & \multicolumn{5}{c}{Qwen-2.5 (32B / 500M)}
 & \multicolumn{5}{c}{Gemma-3 (27B / 270M)}
 & \multicolumn{5}{c}{Llama-3 (70B / 8B)} \\
\cmidrule(lr){2-6} \cmidrule(lr){7-11} \cmidrule(lr){12-16}
 & Writing & Coding & Translation & Math (E) & Math (H)
 & Writing & Coding & Translation & Math (E) & Math (H)
 & Writing & Coding & Translation & Math (E) & Math (H) \\
\midrule
Khisti, delayed expansion
 & 2.56 & 5.21 & 3.22 & 7.04 & 6.81
 & 2.01 & 3.51 & 2.69 & 5.72 & 5.53
 & 5.59 & 7.48 & 7.24 & 7.06 & 6.85 \\
Khisti
 & 2.78 & 5.16 & 3.50 & 6.59 & 6.50
 & 1.21 & 1.77 & 3.09 & 2.26 & 1.92
 & 5.26 & 6.64 & 6.63 & 6.57 & 6.35 \\
\midrule
NaiveTree, delayed expansion
 & 2.41 & 5.31 & 3.42 & 7.06 & 6.49
 & 1.89 & 2.75 & 2.82 & 4.62 & 5.00
 & 5.15 & 7.36 & 6.98 & 6.75 & 6.10 \\
NaiveTree
 & 2.62 & 5.14 & 3.63 & 7.00 & 6.60
 & 1.17 & 1.78 & 3.25 & 2.19 & 1.68
 & 5.18 & 7.19 & 7.01 & 6.88 & 6.24 \\
Naive
 & 2.18 & 4.91 & 2.73 & 6.91 & 6.30
 & 1.74 & 2.60 & 2.80 & 4.91 & 5.17
 & 4.08 & 6.33 & 6.12 & 5.85 & 5.22 \\
\midrule
NSS, delayed expansion
 & 2.00 & 4.48 & 3.12 & 6.24 & 5.74
 & 1.79 & 2.71 & 2.75 & 4.69 & 4.75
 & 4.45 & 7.01 & 6.68 & 5.61 & 5.11 \\
NSS
 & 2.06 & 4.40 & 3.46 & 6.34 & 5.93
 & 1.17 & 1.81 & 3.28 & 2.04 & 1.63
 & 4.45 & 6.86 & 6.69 & 5.69 & 4.93 \\
\midrule
SpecInfer, delayed expansion
 & 2.64 & 5.39 & 3.29 & 7.20 & 6.61
 & 2.00 & 2.84 & 2.74 & 4.66 & 4.87
 & 5.39 & 7.57 & 6.99 & 6.81 & 6.43 \\
SpecInfer
 & 2.79 & 5.33 & 3.67 & 7.12 & 6.76
 & 1.20 & 1.80 & 3.27 & 2.17 & 1.90
 & 5.39 & 7.31 & 7.00 & 6.84 & 6.21 \\
\midrule
SpecTr, delayed expansion
 & 2.69 & 5.32 & 3.37 & 6.87 & 6.59
 & 2.02 & 2.73 & 2.77 & 4.64 & 5.02
 & 5.43 & 7.58 & 7.04 & 7.08 & 6.58 \\
SpecTr
 & 2.85 & 5.33 & 3.69 & 7.04 & 6.65
 & 1.20 & 1.84 & 3.30 & 2.16 & 1.77
 & 5.40 & 7.36 & 7.07 & 7.08 & 6.45 \\
\midrule
BV
 & 2.31 & 4.48 & 2.78 & 6.02 & 5.59
 & 1.79 & 2.45 & 2.76 & 4.76 & 4.73
 & 4.18 & 5.93 & 5.78 & 5.71 & 5.25 \\
\midrule
Traversal, $K$=2
 & 2.70 & 5.09 & 3.22 & 6.63 & 6.20
 & 2.00 & 2.67 & 3.02 & 4.64 & 5.17
 & 4.90 & 6.71 & 6.48 & 6.59 & 6.17 \\
Traversal, $K$=3
 & 2.83 & 5.38 & 3.60 & 7.10 & 6.51
 & 2.08 & 2.84 & 3.20 & 4.96 & 5.52
 & 5.40 & 7.10 & 6.78 & 6.57 & 6.31 \\
Traversal, $K$=4
 & 2.99 & 5.80 & 3.83 & 7.24 & 6.78
 & 2.12 & 3.02 & 3.29 & 5.05 & 5.56
 & 5.62 & 7.36 & 7.07 & 6.96 & 6.86 \\
\bottomrule
\end{tabular}
\caption{Block efficiency by dataset and algorithm}
\label{tab:block_efficiency_by_dataset_and_algorithm}
\end{table}

\begin{table}[t]
\scriptsize
\centering
\setlength{\tabcolsep}{5pt}
\begin{tabular}{lrrrrrr rr}
\toprule
 & \multicolumn{6}{c}{Temperature (top-p=1)} & \multicolumn{2}{c}{Top-p} \\
\cmidrule(lr){2-7} \cmidrule(lr){8-9}
 & 0.2 & 0.4 & 0.6 & 0.8 & 1.0 & 1.2 & 0.90 & 0.99 \\
\midrule
Khisti, delayed expansion
 & 19.59 & 19.94 & 20.06 & 19.42 & 17.36 & 13.58 & 19.50 & 17.80 \\
Khisti
 & 18.23 & 18.66 & 18.90 & 19.07 & 19.37 & 18.53 & 18.11 & 18.47 \\
\midrule
NaiveTree, delayed expansion
 & 22.17 & 22.41 & 22.52 & 22.01 & 21.47 & 19.28 & 21.88 & 21.65 \\
NaiveTree
 & 19.75 & 20.20 & 19.82 & 19.52 & 18.98 & 16.90 & 19.03 & 19.06 \\
Naive
 & 15.40 & 15.86 & 16.58 & 16.85 & 16.74 & 16.05 & 17.51 & 17.39 \\
\midrule
NSS, delayed expansion
 & 22.56 & 22.60 & 21.88 & 21.07 & 19.57 & 15.93 & 19.21 & 19.49 \\
NSS
 & 19.31 & 18.97 & 18.55 & 17.76 & 15.93 & 12.72 & 15.77 & 16.04 \\
\midrule
SpecInfer, delayed expansion
 & 22.59 & 22.67 & 22.68 & 22.94 & 22.82 & 21.06 & 22.69 & 22.87 \\
SpecInfer
 & 19.81 & 19.99 & 20.17 & 20.07 & 19.61 & 17.67 & 19.72 & 19.62 \\
\midrule
SpecTr, delayed expansion
 & 21.06 & 21.88 & 22.15 & 22.47 & 22.35 & 20.21 & 21.89 & 21.93 \\
SpecTr
 & 18.70 & 19.49 & 19.72 & 19.25 & 19.14 & 17.19 & 19.15 & 19.46 \\
\midrule
BV
 & 16.07 & 17.24 & 17.64 & 17.81 & 17.90 & 15.91 & 17.96 & 17.84 \\
\midrule
Traversal, $K$=2
 & 18.93 & 19.61 & 20.14 & 20.20 & 19.83 & 17.99 & 19.68 & 19.65 \\
Traversal, $K$=3
 & 20.14 & 20.78 & 20.88 & 21.18 & 20.87 & 18.68 & 20.93 & 20.90 \\
Traversal, $K$=4
 & 21.09 & 21.56 & 21.90 & 21.63 & 21.52 & 19.79 & 21.68 & 21.60 \\
\bottomrule
\end{tabular}
\caption{Throughput (tokens per second) for Qwen}
\label{tab:tps_qwen}
\end{table}

\begin{table}[t]
\scriptsize
\centering
\setlength{\tabcolsep}{6.5pt}
\begin{tabular}{lrrrrrr rr}
\toprule
 & \multicolumn{6}{c}{Temperature (top-p=1)} & \multicolumn{2}{c}{Top-p} \\
\cmidrule(lr){2-7} \cmidrule(lr){8-9}
 & 0.2 & 0.4 & 0.6 & 0.8 & 1.0 & 1.2 & 0.90 & 0.99 \\
\midrule
Khisti, delayed expansion
 & 4.85 & 5.02 & 5.21 & 5.42 & 5.28 & 3.91 & 4.88 & 5.18 \\
Khisti
 & 5.13 & 5.25 & 5.37 & 5.28 & 4.51 & 3.40 & 5.09 & 5.21 \\
\midrule
NaiveTree, delayed expansion
 & 5.15 & 5.14 & 5.25 & 5.02 & 4.86 & 4.26 & 4.95 & 4.89 \\
NaiveTree
 & 5.16 & 5.20 & 5.17 & 5.11 & 4.97 & 4.44 & 4.98 & 4.96 \\
Naive
 & 4.48 & 4.65 & 4.84 & 4.93 & 4.61 & 4.18 & 4.59 & 4.56 \\
\midrule
NSS, delayed expansion
 & 5.01 & 5.01 & 4.73 & 4.47 & 4.05 & 3.09 & 4.05 & 4.09 \\
NSS
 & 5.05 & 4.98 & 4.89 & 4.65 & 4.19 & 3.38 & 4.16 & 4.22 \\
\midrule
SpecInfer, delayed expansion
 & 5.11 & 5.12 & 5.11 & 5.13 & 5.13 & 4.53 & 5.01 & 5.06 \\
SpecInfer
 & 5.18 & 5.23 & 5.27 & 5.27 & 5.15 & 4.64 & 5.17 & 5.15 \\
\midrule
SpecTr, delayed expansion
 & 4.79 & 5.12 & 5.09 & 5.20 & 5.19 & 4.40 & 4.96 & 4.99 \\
SpecTr
 & 5.04 & 5.24 & 5.28 & 5.20 & 5.15 & 4.62 & 5.14 & 5.21 \\
\midrule
BV
 & 3.96 & 4.20 & 4.30 & 4.36 & 4.37 & 3.93 & 4.39 & 4.38 \\
\midrule
Traversal, $K$=2
 & 4.65 & 4.78 & 4.89 & 4.93 & 4.83 & 4.40 & 4.84 & 4.82 \\
Traversal, $K$=3
 & 4.98 & 5.13 & 5.17 & 5.28 & 5.17 & 4.62 & 5.17 & 5.15 \\
Traversal, $K$=4
 & 5.25 & 5.38 & 5.40 & 5.43 & 5.40 & 4.97 & 5.40 & 5.40 \\
\bottomrule
\end{tabular}
\caption{Block Efficiencies for Qwen}
\label{tab:block_efficiencies_qwen}
\end{table}

\begin{table}[t]
\scriptsize
\centering
\setlength{\tabcolsep}{5pt}
\begin{tabular}{lrrrrrr rr}
\toprule
 & \multicolumn{6}{c}{Temperature (top-p=1)} & \multicolumn{2}{c}{Top-p} \\
\cmidrule(lr){2-7} \cmidrule(lr){8-9}
 & 0.2 & 0.4 & 0.6 & 0.8 & 1.0 & 1.2 & 0.90 & 0.99 \\
\midrule
Khisti, delayed expansion
 & 12.73 & 12.52 & 13.36 & 12.76 & 12.96 & 12.76 & 12.79 & 12.87 \\
Khisti
 & 5.95 & 5.87 & 5.86 & 5.82 & 6.01 & 6.00 & 5.98 & 5.95 \\
\midrule
NaiveTree, delayed expansion
 & 13.36 & 12.97 & 12.55 & 12.53 & 12.75 & 12.93 & 12.94 & 12.82 \\
NaiveTree
 & 6.03 & 5.96 & 5.86 & 6.01 & 6.10 & 5.94 & 6.01 & 6.11 \\
Naive
 & 8.99 & 9.19 & 9.37 & 9.45 & 9.06 & 9.04 & 9.25 & 9.29 \\
\midrule
NSS, delayed expansion
 & 12.72 & 12.38 & 12.54 & 12.47 & 12.04 & 12.40 & 12.05 & 12.64 \\
NSS
 & 5.95 & 5.96 & 6.31 & 5.84 & 5.83 & 5.88 & 5.88 & 5.86 \\
\midrule
SpecInfer, delayed expansion
 & 13.58 & 13.47 & 13.25 & 13.06 & 13.04 & 13.25 & 13.29 & 13.11 \\
SpecInfer
 & 6.13 & 6.20 & 6.21 & 6.09 & 6.35 & 6.24 & 6.27 & 6.21 \\
\midrule
SpecTr, delayed expansion
 & 13.74 & 12.99 & 13.36 & 13.16 & 12.80 & 13.05 & 13.55 & 13.13 \\
SpecTr
 & 5.72 & 5.91 & 5.77 & 6.44 & 6.25 & 6.05 & 6.21 & 6.34 \\
\midrule
BV
 & 10.44 & 10.34 & 10.80 & 10.84 & 10.32 & 10.92 & 10.43 & 10.49 \\
\midrule
Traversal, $K$=2
 & 10.99 & 10.96 & 10.71 & 10.53 & 10.80 & 10.40 & 10.74 & 10.81 \\
Traversal, $K$=3
 & 11.44 & 11.09 & 12.03 & 11.11 & 11.31 & 11.88 & 11.27 & 11.32 \\
Traversal, $K$=4
 & 12.08 & 12.01 & 11.75 & 11.49 & 11.16 & 11.64 & 11.16 & 11.22 \\
\bottomrule
\end{tabular}
\caption{Throughput (tokens per second) for Gemma}
\label{tab:tps_gemma}
\end{table}

\begin{table}[t]
\scriptsize
\centering
\setlength{\tabcolsep}{6.5pt}
\begin{tabular}{lrrrrrr rr}
\toprule
 & \multicolumn{6}{c}{Temperature (top-p=1)} & \multicolumn{2}{c}{Top-p} \\
\cmidrule(lr){2-7} \cmidrule(lr){8-9}
 & 0.2 & 0.4 & 0.6 & 0.8 & 1.0 & 1.2 & 0.90 & 0.99 \\
\midrule
Khisti, delayed expansion
 & 3.82 & 3.78 & 4.06 & 3.90 & 3.91 & 3.89 & 3.86 & 3.92 \\
Khisti
 & 2.05 & 2.04 & 2.02 & 2.01 & 2.07 & 2.06 & 2.09 & 2.06 \\
\midrule
NaiveTree, delayed expansion
 & 3.51 & 3.41 & 3.34 & 3.34 & 3.43 & 3.51 & 3.37 & 3.43 \\
NaiveTree
 & 2.02 & 2.00 & 1.97 & 2.02 & 2.04 & 1.98 & 2.04 & 2.04 \\
Naive
 & 3.47 & 3.34 & 3.40 & 3.53 & 3.50 & 3.57 & 3.37 & 3.37 \\
\midrule
NSS, delayed expansion
 & 3.50 & 3.40 & 3.38 & 3.38 & 3.18 & 3.26 & 3.25 & 3.37 \\
NSS
 & 2.01 & 1.98 & 2.09 & 1.95 & 1.97 & 1.95 & 1.97 & 1.97 \\
\midrule
SpecInfer, delayed expansion
 & 3.44 & 3.40 & 3.42 & 3.45 & 3.42 & 3.51 & 3.36 & 3.38 \\
SpecInfer
 & 2.05 & 2.06 & 2.06 & 2.05 & 2.10 & 2.08 & 2.08 & 2.08 \\
\midrule
SpecTr, delayed expansion
 & 3.47 & 3.35 & 3.50 & 3.44 & 3.39 & 3.56 & 3.42 & 3.37 \\
SpecTr
 & 1.94 & 1.99 & 1.94 & 2.16 & 2.11 & 2.06 & 2.11 & 2.11 \\
\midrule
BV
 & 3.27 & 3.27 & 3.37 & 3.37 & 3.24 & 3.40 & 3.24 & 3.24 \\
\midrule
Traversal, $K$=2
 & 3.58 & 3.55 & 3.46 & 3.49 & 3.49 & 3.45 & 3.49 & 3.49 \\
Traversal, $K$=3
 & 3.71 & 3.62 & 3.93 & 3.70 & 3.68 & 3.93 & 3.68 & 3.68 \\
Traversal, $K$=4
 & 3.96 & 3.97 & 3.83 & 3.82 & 3.65 & 3.90 & 3.67 & 3.67 \\
\bottomrule
\end{tabular}
\caption{Block Efficiencies for Gemma}
\label{tab:block_efficiencies_gemma}
\end{table}

\begin{table}[t]
\scriptsize
\centering
\setlength{\tabcolsep}{5pt}
\begin{tabular}{lrrrrrr rr}
\toprule
 & \multicolumn{6}{c}{Temperature (top-p=1)} & \multicolumn{2}{c}{Top-p} \\
\cmidrule(lr){2-7} \cmidrule(lr){8-9}
 & 0.2 & 0.4 & 0.6 & 0.8 & 1.0 & 1.2 & 0.90 & 0.99 \\
\midrule
Khisti, delayed expansion
 & 13.20 & 13.14 & 13.56 & 13.44 & 13.52 & 11.44 & 13.91 & 13.67 \\
Khisti
 & 13.52 & 13.46 & 13.82 & 13.89 & 13.79 & 12.75 & 14.06 & 13.99 \\
\midrule
NaiveTree, delayed expansion
 & 13.79 & 14.11 & 13.99 & 14.13 & 14.24 & 13.56 & 14.05 & 14.22 \\
NaiveTree
 & 14.22 & 14.39 & 14.41 & 15.32 & 14.50 & 14.34 & 15.35 & 15.12 \\
Naive
 & 10.61 & 11.09 & 11.57 & 11.96 & 12.08 & 11.38 & 12.06 & 11.87 \\
\midrule
NSS, delayed expansion
 & 13.44 & 13.43 & 13.31 & 12.77 & 11.90 & 11.01 & 11.69 & 12.05 \\
NSS
 & 13.77 & 13.36 & 13.36 & 13.57 & 11.93 & 12.01 & 12.69 & 12.59 \\
\midrule
SpecInfer, delayed expansion
 & 13.88 & 14.10 & 14.50 & 14.48 & 14.51 & 14.05 & 14.24 & 14.38 \\
SpecInfer
 & 13.97 & 14.30 & 14.59 & 15.65 & 14.74 & 15.05 & 15.12 & 15.42 \\
\midrule
SpecTr, delayed expansion
 & 13.92 & 13.94 & 14.57 & 14.60 & 14.59 & 13.47 & 14.18 & 14.55 \\
SpecTr
 & 14.11 & 14.27 & 14.30 & 15.33 & 14.68 & 15.22 & 15.45 & 15.54 \\
\midrule
BV
 & 10.97 & 11.63 & 11.98 & 12.55 & 12.79 & 12.24 & 12.74 & 12.70 \\
\midrule
Traversal, $K$=2
 & 12.91 & 12.83 & 13.87 & 13.69 & 14.01 & 14.03 & 14.19 & 14.01 \\
Traversal, $K$=3
 & 13.62 & 14.02 & 13.83 & 14.17 & 14.18 & 14.31 & 14.25 & 14.27 \\
Traversal, $K$=4
 & 13.97 & 14.52 & 15.04 & 15.01 & 14.77 & 15.07 & 14.99 & 14.79 \\
\bottomrule
\end{tabular}
\caption{Throughput (tokens per second) for Llama}
\label{tab:tps_llama}
\end{table}

\begin{table}[t]
\scriptsize
\centering
\setlength{\tabcolsep}{6.5pt}
\begin{tabular}{lrrrrrr rr}
\toprule
 & \multicolumn{6}{c}{Temperature (top-p=1)} & \multicolumn{2}{c}{Top-p} \\
\cmidrule(lr){2-7} \cmidrule(lr){8-9}
 & 0.2 & 0.4 & 0.6 & 0.8 & 1.0 & 1.2 & 0.90 & 0.99 \\
\midrule
Khisti, delayed expansion
 & 6.49 & 6.52 & 6.77 & 6.97 & 7.32 & 6.59 & 6.88 & 7.19 \\
Khisti
 & 6.58 & 6.58 & 6.75 & 6.85 & 5.72 & 4.25 & 6.78 & 6.81 \\
\midrule
NaiveTree, delayed expansion
 & 6.34 & 6.54 & 6.46 & 6.50 & 6.60 & 6.27 & 6.50 & 6.55 \\
NaiveTree
 & 6.42 & 6.51 & 6.51 & 6.61 & 6.56 & 6.20 & 6.65 & 6.52 \\
Naive
 & 5.06 & 5.30 & 5.54 & 5.70 & 5.73 & 5.41 & 5.76 & 5.68 \\
\midrule
NSS, delayed expansion
 & 6.24 & 6.24 & 6.21 & 5.94 & 5.49 & 5.06 & 5.46 & 5.56 \\
NSS
 & 6.26 & 6.05 & 6.06 & 5.82 & 5.41 & 5.20 & 5.49 & 5.48 \\
\midrule
SpecInfer, delayed expansion
 & 6.42 & 6.60 & 6.71 & 6.72 & 6.76 & 6.53 & 6.68 & 6.69 \\
SpecInfer
 & 6.35 & 6.46 & 6.60 & 6.71 & 6.63 & 6.45 & 6.53 & 6.66 \\
\midrule
SpecTr, delayed expansion
 & 6.61 & 6.58 & 7.07 & 6.91 & 6.88 & 6.35 & 6.69 & 6.87 \\
SpecTr
 & 6.50 & 6.56 & 6.60 & 6.69 & 6.76 & 6.66 & 6.78 & 6.82 \\
\midrule
BV
 & 4.81 & 5.11 & 5.29 & 5.54 & 5.60 & 5.39 & 5.64 & 5.59 \\
\midrule
Traversal, $K$=2
 & 5.83 & 5.79 & 6.23 & 6.19 & 6.29 & 6.30 & 6.41 & 6.31 \\
Traversal, $K$=3
 & 6.21 & 6.37 & 6.39 & 6.47 & 6.50 & 6.51 & 6.49 & 6.50 \\
Traversal, $K$=4
 & 6.40 & 6.70 & 6.86 & 6.84 & 6.80 & 6.90 & 6.88 & 6.83 \\
\bottomrule
\end{tabular}
\caption{Block Efficiencies for Llama}
\label{tab:block_efficiencies_llama}
\end{table}

\end{document}